\documentclass[twoside]{article}
\usepackage[accepted]{aistats_arxiv}
\usepackage{svg}

\usepackage{amssymb}
\usepackage{amsthm}
\usepackage{enumitem}
\usepackage{subcaption}
\usepackage{verbatim}
\usepackage{xcolor}
\usepackage{graphicx}
\usepackage{enumitem}
\usepackage{subfiles}
\usepackage{amsmath}
\usepackage{todonotes}
\usepackage{tikz-qtree,tikz-qtree-compat}
\usepackage{scrextend}
\usepackage{framed}
\usepackage{placeins}
\usepackage{setspace}
\usepackage{mathtools,collcell,eqparbox}
\usepackage{ dsfont }
\usepackage{multicol}
\usepackage{wrapfig}
\usepackage{pgfplots}
\usepackage{tikz}
\usepackage{listings}
\usepackage{courier}
\usepackage[round]{natbib}
\usepackage{url}
\usepackage{color}
\usepackage[title]{appendix}
\usepackage{algorithmic}
\usepackage{algorithm}

\newcommand{\umax}{\|u\|_{L^\infty(\mu_{s,a})}}

\usepackage[pagebackref = true]{hyperref}
\hypersetup{
    colorlinks = true,
    linkcolor = red,
    anchorcolor = blue,
    citecolor = blue,
    filecolor = blue,
    pagecolor=blue,
    urlcolor = blue
    }

\newcounter{BMatrix}

\newcommand{\setmaxwd}[1]{%
  \eqmakebox[BM-\theBMatrix][\BMalign]{$#1$}%
}
\MHInternalSyntaxOn

\MHInternalSyntaxOff
\makeatother

\newtheorem{theorem}{Theorem}[section]
\newtheorem{proposition}[theorem]{Proposition}
\newtheorem{assumption}{Assumption}

\newtheorem{lemma}[theorem]{Lemma}
\theoremstyle{definition}
\newtheorem{definition}{Definition}
\theoremstyle{remark}
\newtheorem*{remark}{Remark}

\DeclareMathOperator*{\esssup}{ess\,sup}
\DeclareMathOperator*{\essinf}{ess\,inf}
\newcommand{\inv}{^{-1}}

\newcommand{\1}{\mathds{1}}

\newcommand{\R}{\mathbb{R}}

\newcommand{\Z}{\mathbb{Z}}

\newcommand{\var}{\text{Var}}

\newcommand{\cov}{\text{Cov}}

\newcommand{\del}{\partial}
\newcommand{\da}{\downarrow}
\newcommand{\ra}{\rightarrow}

\newcommand{\cd}{\cdot}
\newcommand{\ds}{\dots}

\newcommand{\cB}{\mathcal{B}}

\newcommand{\cF}{\mathcal{F}}
\newcommand{\cG}{\mathcal{G}}
\newcommand{\cH}{\mathcal{H}}

\newcommand{\cK}{\mathcal{K}}
\newcommand{\cL}{\mathcal{L}}
\newcommand{\cM}{\mathcal{M}}

\newcommand{\cP}{\mathcal{P}}

\newcommand{\cR}{\mathcal{R}}
\newcommand{\cS}{\mathcal{S}}
\newcommand{\cT}{\mathcal{T}}

\newcommand{\unif}{\text{Unif}}

\newcommand{\set}[1]{\left\{{#1}\right\}}

\newcommand{\norm}[1]{\left\|#1\right\|}

\newcommand{\abs}[1]{\left|#1\right|}
\newcommand{\sqbk}[1]{\left[ #1 \right]}
\newcommand{\sqbkcond}[2]{\left[ #1 \middle| #2 \right]}

\newcommand{\crbk}[1]{\left( #1 \right)}

\newcommand{\argmax}[1]{\underset{#1}{\operatorname{arg}\,\operatorname{max}}\;}

\def \cF {\mathcal{F}}

\def \cG {\mathcal{G}}
\def \cM {\mathcal{M}}
\def \cB {\mathcal{B}}

\def \var {\mathsf{Var}}

\def\EE{\mathbb{E}}

\def\PP{\mathbb{P}}

\def\RR{\mathbb{R}}

\newcommand{\calM}{{\mathcal{M}}}

\newcommand{\calP}{{\mathcal{P}}}

\newcommand{\calR}{{\mathcal{R}}}

\newcommand{\calT}{{\mathcal{T}}}

\usepackage{caption}

\begin{document}

%

%

\twocolumn[
\runningtitle{Sample Complexity of DR Q-learning}
\aistatstitle{A Finite Sample Complexity Bound\\ for Distributionally Robust $Q$-learning}

\aistatsauthor{ Shengbo Wang \And Nian Si \And  Jose Blanchet \And Zhengyuan Zhou }

\aistatsaddress{ Stanford University \And  The University of Chicago \And Stanford University \And New York University} ]

\begin{abstract}
We consider a reinforcement learning setting in which the deployment environment is different from the training environment. Applying a robust Markov decision processes formulation, we extend the distributionally robust $Q$-learning framework studied in \cite{pmlr-v162-liu22a}. Further, we improve the design and analysis of their multi-level Monte Carlo estimator. Assuming access to a simulator, we prove that the worst-case expected sample complexity of our algorithm to learn the optimal robust $Q$-function within an $\epsilon$ error in the sup norm is upper bounded by $\tilde O(|S||A|(1-\gamma)^{-5}\epsilon^{-2}p_{\wedge}^{-6}\delta^{-4})$, where $\gamma$ is the discount rate, $p_{\wedge}$ is the non-zero minimal support probability of the transition kernels and $\delta$ is the uncertainty size. This is the first sample complexity result for the model-free robust RL problem. Simulation studies further validate our theoretical results.

\end{abstract}
\section{Introduction}
Reinforcement learning (RL)~\citep{powell2011approDP, Bertsekas2012control, Szepesvari2012rlAlgo, sutton2018reinforcement} 
has witnessed impressive empirical success in simulated environments, with applications spanning domains such as robotics~\citep{Kober2013RLinRoboticsSurvey, Sergey2017DeepRLinRobotics}, computer vision \citep{Sergey2017RLsingleImage, huang2017RLinCV}, finance \citep{li2009americanOption,CHOI2009rlSavingBehavior,Deng2017rlFinanceSignal} and achieving superhuman performance in well-known games such as Go and poker~\citep{alphago2016, alphazero2018}.

However, existing RL algorithms often 
make the implicit assumption that the training environment (i.e. a simulator) is the same as the deploying environment, thereby rendering the learned policy \textit{fragile}. This fragility presents a significant impediment for carrying the remarkable success of RL into real environments, because in practice, such discrepancy between training and deploying environments is ubiquitous. 
On the one hand, simulator models often cannot capture the full complexity of the real environment, and hence will be mis-specified.
On the other hand, even if a policy is trained directly in a real environment, the new deployment environment may not be the same and hence suffer from distributional shifts.

As an example of the latter, 
personalized promotions engine (learned from existing user browsing data collected in one region or market)  may need to be deployed in a different region when the company intends to enter a new market.
The new market may have similar but different population characteristics.
Another example occurs in robotics, where, as articulated in~\cite{pmlr-v130-zhou21d} ``a robot trained to perform certain maneuvers (such as walking~\cite{Abbeel2013robotWalking} or folding laundry~\citep{Abbeel2010robotFold}) in an environment can fail catastrophically~\citep{Abbeel2015DARPA} in a slightly different environment, where the terrain landscape (in walking) is slightly altered or the laundry object (in laundry folding) is positioned differently".

\par Motivated by the necessity of policy robustness in RL applications, \cite{pmlr-v130-zhou21d} adapted the distributionally robust (DR) Markov decision processes (MDPs) to a tabular RL setting and proposed a DR-RL paradigm. Subsequent works \cite{yang2021} have improved on the sample complexity bounds, although the optimal bound is still unknown as of this writing. However, all these works all adopt a model-based approach which, as widely known, is computationally intensive, requires extensive memory storage, and does not generalize to function approximation settings. Motivated by this concern, the very recent work~\cite{pmlr-v162-liu22a} introduced the first distributionally robust $Q$-learning for robust MDPs, thus showing that $Q$-learning can indeed be made distributionally robust. However, an important issue is that the expected number of samples needed to run the algorithm in ~\cite{pmlr-v162-liu22a} to converge to a fixed error distributionally robust optimal policy is infinite. As such, this naturally motivates the following question:

\textit{Can we design a distributionally robust $Q$-learning that has finite sample complexity guarantee?}

\subsection{Our Contributions}
In this paper, we extend the MLMC-based distributionally robust Bellman estimator in \cite{pmlr-v162-liu22a} such that the expected sample size of constructing our estimator is of \textit{constant order}. We establish unbiasedness and moment bounds for our estimator in Propositions \ref{prop:unbiased} and \ref{prop:var_sup_bound} that are essential to the complexity analysis. Hinging on these properties, we prove that the expected sample complexity of our algorithm is $\tilde O\crbk{|S||A|(1-\gamma)^{-5}\epsilon^{-2}p_{\wedge}^{-6}\delta^{-4}}$ under rescaled-linear or constant stepsizes, where $|S|$ and $|A|$ are the number of states and actions, $\gamma\in(0,1)$ the discount factor,  $\epsilon$ the target error in the infinity norm of the DR $Q$-function, $p_{\wedge}$ the minimal support probability, and $\delta$ the size of the (see Theorems \ref{thm:algo_error_bound} and \ref{thm:sample_complexity}). Our result is based on the finite sample analysis of stochastic approximations (SA) framework recently established by \cite{chen2020}. To our knowledge, this is the first model-free algorithm and analysis that guarantee solving the DR-RL problem with a finite expected sample complexity. Further, our complexity is tight in $|S||A|$ and nearly tight in the effective horizon $(1-\gamma)\inv$ at the same time. Finally, we numerically exhibit the validity of our theorem predictions and demonstrate the improvements  of our algorithm over that in \cite{pmlr-v162-liu22a}. 

\subsection{Related Work}

Distributionally robust optimization (DRO) is well-studied in the supervised learning setting; see, e.g., \cite{bertsimas2004price,delage2010distributionally,Hu2012KullbackLeiblerDC,shafieezadeh-abadeh_distributionally_2015,bayraksan2015data,gao2016distributionally,namkoong2016stochastic,duchi2016statistics,staib2017distributionally,shapiro2017distributionally,lam2017empirical,volpi2018generalizing,raginsky_lee,nguyen2018distributionally,yang2020Wasserstein,MohajerinEsfahani2017,zhao2017distributionally,abadeh2018Wasserstein,ZHAO2018262,sinha2018certifiable,gao2018robust,wolfram2018,ghosh2019robust,blanchet2016quantifying,duchi2018learning,lam2019recovering,duchi2019distributionally,ho2020distributionally}. Those works focus on the optimization formulation, algorithms, and statistical properties in settings where labeled data and a pre-specified loss are available. In those settings, vanilla empirical risk minimizers are outperformed by distributionally robust solutions because of either overfitting or distributional shifts.

In recent years, distributionally robust formulations also find applications in a wide range of research areas including dimensionality reduction under fairness \citep{vu2022distributionally} and model selection \citep{cisneros2020distributionally}. 

Minimax sample complexities of standard tabular RL have been studied extensively in recent years. \cite{azar2013,sidford2018near_opt,pmlr-v125-agarwal20b,li2020breaking} proposed algorithms and proved optimal upper bounds (the matching lower bound is proved in \cite{azar2013}) $\tilde \Theta(|S||A|(1-\gamma)^{-3}\epsilon^{-2})$ of the sample complexity to achieve $\epsilon$ error in the model-based tabular RL setting. The complexity of model-free $Q$-learning has also been studied extensively \citep{even2003learning,wainwright2019l_infty,li2021q}. It has been shown by \cite{li2021q} to have a minimax sample complexity $\tilde \Theta(|S||A|(1-\gamma)^{-4}\epsilon^{-2})$. Nevertheless, variance-reduced variants of the $Q$-learning, e.g., \cite{wainwright2019}, achieves the aforementioned model-based sample complexity lower bound  $\tilde \Theta(|S||A|(1-\gamma)^{-3}\epsilon^{-2})$. 

Recent advances in sample complexity theory of $Q$-learning and its variants are propelled by the breakthroughs in finite time analysis of SA. \cite{wainwright2019l_infty} proved a sample path bound for the SA recursion. This enables variance reduction techniques that help to achieve optimal learning rate in \cite{wainwright2019}. In comparison, \cite{chen2020} established finite sample guarantees of SA only under a second moment bound on the martingale difference noise sequence.

Our work uses the theoretical framework of the classical minimax control and robust MDPs; see, e.g., \citet{gonzalez-trejo2002,Iyengar2005, wiesemann2013,huan2010,shapiro2022}, where those works established the concept of distributional robustness in MDPs and derive the distributionally robust Bellman equation. 
\par Recently, learning distributionally robust policies from data gains attention \citep{si2020distributional,pmlr-v130-zhou21d, yang2021,pmlr-v162-liu22a,Shi2022ModelBase}. Among those works, \cite{si2020distributional} studied the contextual bandit setting. \cite{pmlr-v130-zhou21d, Panaganti2021,yang2021, Shi2022ModelBase} focused on the model-based tabular RL setting; \cite{pmlr-v162-liu22a} tackled the DR-RL problem using a model free approach. 

\section{Distributionally Robust Policy Learning Paradigm}

\subsection{Standard Policy Learning}
Let $\calM_0 = \left(S, A, R, \calP_0, \calR_0, \gamma\right)$ be an MDP, where $ S$, $ A$, and $R\subsetneq \R_{\geq 0}$ are finite state, action, and reward spaces respectively. 
$\calP_0 = \set{p_{s,a}, s\in S,a\in A}$ and $
    \cR_0  = \set{\nu_{s,a},s\in S,a\in A}$ 
are the sets of the reward and transition distributions. $\gamma \in (0,1)$ is the discount factor. Define $r_{\max} = \max\set {r\in R}$ the maximum reward. We assume that the transition is Markovian, i.e., at each state $s\in  S$, if action $a\in A$ is chosen, then the subsequent state is determined by the conditional distribution $p_{s,a}(\cdot)= p(\cdot|s, a)$. 
The decision maker will therefore receive a randomized reward $r\sim \nu_{s,a}$. 
Let $\Pi$ be the history-dependent policy class (see  Section 1 of the supplemental materials for a rigorous construction). For $\pi\in \Pi$, the value function $V^{\pi}(s)$ is defined as:
$$V^{\pi}(s) := \EE\sqbkcond{\sum_{t=0}^\infty \gamma^{t}r_t }{s_0 = s}.$$ The optimal value function $
V^*(s) \coloneqq \max_{\pi\in\Pi} V^\pi(s)$, 
$\forall s\in S$. It is well known that the optimal value function is the unique solution of the following Bellman equation:
\begin{equation*}
V^*(s) = \max_{a\in A} \left\lbrace\EE_{r\sim \nu_{s,a}}[r] + \gamma\EE_{s'\sim p_{s,a}}\left[V^*(s')\right]\right\rbrace.
\end{equation*}  
An important implication of the Bellman equation is that it suffices to optimize within the stationary Markovian deterministic policy class. 
\par The optimal $Q$-function and its Bellman equation:
\begin{equation*}
\begin{aligned}
    Q^*(s,a) &:= \EE_{r\sim \nu_{s,a}}[r] + \gamma\EE_{s'\sim p_{s,a}}\left[V^*(s')\right]\\
    &= \EE_{r\sim \nu_{s,a}}[r] + \gamma\EE_{s'\sim p_{s,a}}\left[\max_{b\in A}Q^*(s',b)\right].
\end{aligned}
\end{equation*}
The optimal policy
$\pi^*(s) = \arg\max_{a\in A}Q^*(s,a)$. Therefore, policy learning in RL environments can be achieved if we can learn a good estimate of $Q^*$.

\subsection{Distributionally Robust Formulation}
We consider a DR-RL setting, where both transition probabilities and rewards are perturbed based on the KL divergence $D_{\textrm{KL}}(P\|Q) := \int_\Omega \log\frac{dP}{dQ}P(d\omega)$ when $P\ll Q$ ($P$ is absolutely continuous w.r.t. $Q$). For each $(s,a)\in S\times A$, we define KL uncertainty set that are centered at $p_{s,a}\in\cP_0$ and $\nu_{s,a}\in\cR_0$ by $
\calP_{s,a}(\delta)\coloneqq \left\lbrace p : D_{\textrm{KL}}\left(p\|p_{s,a}\right)\leq \delta\right\rbrace$ and $\calR_{s,a}(\delta)\coloneqq \left\lbrace \nu: D_{\textrm{KL}}(\nu\|\nu_{s,a})\leq\delta\right\rbrace$.
The parameter $\delta > 0$ controls the size of the uncertainty sets. These uncertainty sets quantify the possible distributional shifts from the reference model $\cP_0,\cR_0$. 
\begin{definition}\label{def:DRvalue}
The DR Bellman operator $\cB_\delta$ for the value function is defined as the mapping
\begin{equation} \label{def.robust_Bellman_opt}
\begin{aligned}
&\cB_\delta(v)(s):=\\
&\max_{a\in A} \inf_{\mbox{$\begin{subarray}{c} p\in\calP_{s,a}(\delta),\\
		\nu\in \calR_{s,a}(\delta)\end{subarray}$}}
\left\lbrace  \EE_{r\sim \nu} [r] + \gamma\EE_{s'\sim p}\left[v(s')\right]\right\rbrace.
\end{aligned}
\end{equation}
Define the optimal DR value function $V^*_\delta$ as the solution of the DR Bellman equation: 
\begin{equation}\label{def.robust_Bellman}
    V^*_\delta =  \cB_\delta(V^*_\delta)
\end{equation}
\end{definition}
\begin{remark}
The definition assumes the existence and uniqueness of a fixed point of the DR Bellman equation. This is a consequence of $\cB_\delta$ being a contraction. Moreover, it turns out that under the notion of \textit{rectangularity} \citep{Iyengar2005, wiesemann2013}, this definition is equivalent to the minimax control optimal value
\[
V^*_{\delta}(s)= \sup_{\pi\in\Pi}\inf_{P\in \cK^\pi(\delta)}\EE_{P}\sqbkcond{\sum_{t=0}^\infty \gamma^t r_t }{s_0 = s}
\]
for some appropriately defined \textit{history-dependent} policy class $\Pi$ and $\pi$-consistent uncertainty set of probability measures $\cK^\pi(\delta)$ on the sample path space; cf. \cite{Iyengar2005}.  Intuitively, this is the optimal value when the controller chooses a policy $\pi$, an adversary observes this policy and chooses a possibly history-dependent sequence of reward and transition measure within some uncertainty set indexed by a parameter $\delta$ that is consistent with this policy. Therefore, we can interpret $\delta > 0$ as the power of this adversary. The equivalence of minimax control optimal value and Definition \ref{def:DRvalue} suggests the optimality of stationary deterministic Markov control policy and, under such policy, stationary Markovian adversarial distribution choice. We will rigorously discuss this equivalence in  Section 1 of the supplemental materials.
\end{remark}

\subsection{Strong Duality}
The r.h.s. of \eqref{def.robust_Bellman_opt} could be hard to work with because the measure underlying the expectations are not directly accessible. To resolve this, we use strong duality:
\begin{lemma} [\cite{Hu2012KullbackLeiblerDC}, Theorem 1] \label{lemma.dual}
Suppose $H(X)$ has finite moment generating function in the neighborhood of zero. Then for any $\delta >0$, 
	\begin{equation*}
	\begin{aligned}
	&\sup_{P: D_{\emph{KL}}(P\|P_0)\leq \delta}\EE_P\left[H(X)\right] =\\
	&\inf_{\alpha\geq 0}\left\lbrace \alpha\log\EE_{P_0}\left[e^{H(X)/\alpha}\right] +\alpha\delta\right\rbrace.
	\end{aligned}
	\end{equation*}
\end{lemma}
Boundedness of $Q$ allow us to directly apply Lemma \ref{lemma.dual} to the r.h.s. of \eqref{def.robust_Bellman}. The DR value function $V^{*}_{\delta}$ in fact satisfies the following \textit{dual form} of the DR Bellman's equation. 
\begin{equation}\label{eq.robust_dual}
\begin{aligned}
&V^{ *}_{\delta}(s) = \\
& \max_{a\in A} \left\lbrace \sup_{\alpha\geq 0}\left\lbrace -\alpha\log\EE_{r\sim \nu_{s,a}}\left[e^{-r/\alpha}\right] - \alpha\delta\right\rbrace \right. + \\
& \left. \gamma\sup_{\beta\geq 0}\left\lbrace -\beta\log\EE_{s'\sim p_{s,a}}\left[e^{-V^{*}_{\delta}(s')/\beta}\right] - \beta\delta\right\rbrace \right\rbrace.
\end{aligned}
\end{equation}

\subsection{Distributionally Robust $Q$-function and its Bellman Equation} 
As in the classical policy learning paradigm, we make use of the optimal DR state-action value function, a.k.a. $Q$-function, for solving the DR control problem. The $Q$-function maps $(s,a)$ pairs to the reals, thence can be identified with $Q\in \R^{S\times A}$. We will henceforth assume this identification. Let us define the notation $v(Q)(s) = \max_{b\in A}Q(s,b)$. We proceed to rigorously define the optimal $Q$-function and its Bellman equation. 
\begin{definition}
The optimal DR $Q$-function is defined as
\begin{equation}\label{eq.def_delta_Q*}
Q_\delta^*(s,a) := \inf_{\mbox{$\begin{subarray}{c} p\in\calP_{s,a}(\delta),\\
		\nu\in \calR_{s,a}(\delta)\end{subarray}$}}
    \left\lbrace  \EE_{r\sim \nu} [r] + \gamma\EE_{s'\sim p}\left[V^*_\delta(s')\right] \right\rbrace
\end{equation}
where $V^*_\delta$ is the DR optimal value function in Definition \ref{def:DRvalue}.
\end{definition}
By analogy with the Bellman operator, we can define the DR Bellman operator for the $Q$-function as follows:
\begin{definition}
    Given $\delta>0$ and $Q\in\RR^{ S\times A}$, the \textit{primal form} of the DR Bellman operator $\calT_\delta:\RR^{ S\times  A}\to \RR^{ S\times  A}$ is defined as
    \begin{equation}\label{eq.def_op_pr}
    \begin{aligned}
    &\calT_\delta(Q)(s,a):= \\
    &\inf_{\mbox{$\begin{subarray}{c} p\in\calP_{s,a}(\delta),\\
		\nu\in \calR_{s,a}(\delta)\end{subarray}$}}
    \left\lbrace  \EE_{r\sim \nu} [r] + \gamma\EE_{s'\sim p}\left[v(Q)(s')\right] \right\rbrace\\
    \end{aligned}
    \end{equation}
    The \textit{dual form} of the DR Bellman operator is defined as
    \begin{equation}\label{eq.def_op}
    \begin{aligned}
    &\calT_\delta(Q)(s,a)\coloneqq\\
    & \sup_{\alpha\geq 0}\left\lbrace -\alpha\log\EE_{r\sim \nu_{s,a}}\left[e^{-r/\alpha}\right] - \alpha\delta\right\rbrace+ \\
    & \gamma\sup_{\beta\geq 0}\left\lbrace -\beta\log\EE_{s'\sim p_{s,a}}\left[e^{-v(Q)(s')/\beta}\right] - \beta\delta\right\rbrace.
    \end{aligned}
    \end{equation}
\end{definition}
The equivalence of the primal and dual form follows from Lemma \ref{lemma.dual}. Note that by definition \eqref{eq.def_delta_Q*} and the Bellman equation \eqref{def.robust_Bellman}, we have $v(Q_{\delta}^*) = V^*_\delta$. So, our definition implies that $Q^{*}_\delta$ is a fixed point of $\calT_\delta$ and the Bellman equation $Q^{*}_\delta = \cT_\delta(Q^{*}_\delta)$.
\par The optimal DR policy can be extracted from the optimal $Q$-function by $\pi_\delta^*(s) = \arg\max_{a\in A}Q_\delta^*(s,a).$
Hence the goal the DR-RL paradigm is to learn the $\delta$-DR $Q$-function and extract the corresponding robust policy. 

\section{$Q$-Learning in Distributionally Robust RL}
\subsection{A Review of Synchronized $Q$-Learning and Stochastic Approximations}\label{section.review_Q}
The synchronized $Q$-learning estimates the optimal $Q$-function using point samples. The classical synchronous $Q$-learning proceeds as follows. At iteration $k\in \Z_{\geq 0}$ and each $(s,a)\in S\times A$, we draw samples $r\sim \nu_{s,a}$ and $s'\sim p_{s,a}$. Then perform the $Q$-learning update
\begin{equation}
\begin{aligned}
&Q_{k+1}(s,a)=\\
&(1-\alpha_{k})Q_k(s,a) + \alpha_{k}(r +\gamma v(Q_k)(s'))
\end{aligned}\label{eq.q-learning}
\end{equation}
for some chosen step-size sequence $\set{\alpha_k}$. 
\par Stochastic approximations (SA) for the fixed point of a contraction operator $\cH$ refers to the class of algorithms using the update
\begin{equation}\label{eq.sa_update}
X_{k+1} = (1-\alpha_k)X_{k} + \alpha_k\cH(X_k) + W_k.
\end{equation}
$\set{W_k}$ is a sequence satisfying $\EE[W_{k}|\cF_{k-1}] = 0$,
thence is known as the \textit{martingale difference noise}. The asymptotics of the above recursion are well understood, cf. \cite{kushner2013stochastic}; while finite time behavior is discussed in the literature review. The recursion representation of the $Q$-learning \eqref{eq.q-learning} fits into the SA framework: Note that $r+\gamma v(Q)(s')$ is an \textit{unbiased} estimator of $\cT(Q)$ where $\cT$ is the Bellman operator for the $Q$-function. This representation motivates the DR $Q$-learning. 

\subsection{Distributionally Robust $Q$-learning}
\par A foundation to the possibility of employing a $Q$-learning is the following result.
\begin{proposition}\label{prop:contraction}
The DR Bellman operator $\cT_\delta$ is a $\gamma$-contraction on the Banach space $(\R^{S\times A},\|\cd\|_\infty)$.
\end{proposition}
\par Given a simulator, a natural estimator for $\cT_\delta(Q)$ is the empirical dual Bellman operator: replace the population transition and reward measures in \eqref{eq.def_op} with the empirical version. However, the nonlinearity in the underlying measure, which can be seen from the dual functional, makes this estimator biased in general. Instead, \cite{pmlr-v162-liu22a} propose an alternative by employing the idea in \cite{ blanchet2019unbiased}; i.e., producing unbiased estimate of nonlinear functional of a probability measure using multi-level randomization. Yet, the number of samples requested in every iteration in \cite{pmlr-v162-liu22a} is infinite in expectation. We improve this by extending the construction to a regime where the expected number of samples used is constant. 

\par Before moving forward, we introduce the following notation. Denote the empirical distribution on $n$ samples with $\nu_{s,a,n}$ and $p_{s,a,n}$ respectively; i.e. for $f:U\ra \R$, where $U$ could be the $S$ or $R$, 
\[
\EE_{u\sim \mu_{s,a,n}}f(u) = \frac{1}{n}\sum_{j=1}^nf(u_i)
\]
for $\mu = \nu,p$ and $u_i = r_i,s'_i$. Moreover, we use $\mu^{O}_{s,a,n}$ and $\mu^{E}_{s,a,n}$ to denote the empirical distribution formed by the odd and even samples in $\mu_{s,a,2n}$. With this notation, we defined our estimator:
\begin{definition}
    For given $g\in(0,1)$ and $Q\in\RR^{ S\times A}$, define the MLMC-DR estimator:
    \begin{equation}
        \widehat{\calT}_{\delta,g}(Q)(s,a)\coloneqq\widehat{R}_\delta(s,a)+\gamma \widehat{V}_\delta(Q)(s,a).
    \end{equation}
    For $\widehat{R}_\delta(s,a)$ and $\widehat{V}_\delta(s,a)$, we sample $N_1,N_2$ from a geometric distribution $\mathrm{Geo}(g)$ independently, i.e.,  $\mathbb{P}(N_j=n)=p_n:=g(1-g)^n, n\in\Z_{\geq 0},j=1,2$. Then, we  draw $2^{N_1+1}$ samples $r_i\sim \nu_{s,a}$ and $2^{N_2+1}$ samples $s'_i\sim p_{s,a}$. Finally, we compute
    \begin{align}
        \widehat{R}_\delta(s,a)&\coloneqq                      r_1+\frac{\Delta_{N_1,\delta}^R}{p_{N_1}},\label{eq.R}.\\
        \widehat{V}_{\delta}(Q)(s,a)&\coloneqq v(Q)(s_1')+\frac{\Delta_{N_2,\delta}^P(Q)}{p_{N_2}}\label{eq.V}.
    \end{align}
    where
    \begin{equation}\label{def.delta_r}
    \begin{aligned}
    &\Delta_{n,\delta}^R\coloneqq\\
        &\sup_{\alpha\geq0}\left\lbrace -\alpha\log\EE_{r\sim\nu_{ s,a,2^{n+1}}}\sqbk{e^{-r/\alpha}} - \alpha\delta\right\rbrace-\\
        &\frac{1}{2}\sup_{\alpha\geq0}\left\lbrace -\alpha\log\EE_{r\sim\nu^{E}_{s,a,2^{n}}}\sqbk{e^{-r/\alpha}} - \alpha\delta\right\rbrace-\\ &\frac{1}{2}\sup_{\alpha\geq0}\left\lbrace -\alpha\log\EE_{r\sim\nu^{O}_{s,a,2^{n}}}\sqbk{e^{-r/\alpha}} - \alpha\delta\right\rbrace
    \end{aligned}      
    \end{equation}and
    \begin{equation}\label{def.delta_q}
    \begin{aligned}
        &\Delta_{n,\delta}^P(Q)\coloneqq\\
        &\sup_{\beta\geq0}\left\lbrace -\beta\log\EE_{s'\sim p_{s,a,2^{n+1}}}\sqbk{e^{-v(Q)(s')/\beta}} - \beta\delta\right\rbrace-\\
        &\frac{1}{2}\sup_{\beta\geq0}\left\lbrace -\beta\log\EE_{s'\sim p^{E}_{s,a,2^{n}}}\sqbk{e^{-v(Q)(s')/\beta}} - \beta\delta\right\rbrace-\\
        &\frac{1}{2}\sup_{\beta\geq0}\left\lbrace -\beta\log\EE_{s'\sim p^{O}_{s,a,2^{n}}}\sqbk{e^{-v(Q)(s')/\beta}} - \beta\delta\right\rbrace.
    \end{aligned}
    \end{equation}  
\end{definition}
Let $\set{\widehat{\calT}_{\delta,g,k};k\in \Z_{\geq 0}}$ be i.i.d. copies of $\widehat{\calT}_{\delta,g}$. We construct our DR $Q$-Learning algorithm  in Algorithm \ref{alg.Q_learning}. 

\begin{algorithm}[ht]
   \caption{Multi-level Monte Carlo Distributionally Robust $Q$-Learning (MLMCDR $Q$-learning)}
   \label{alg.Q_learning}
\begin{algorithmic}
   \STATE {\bfseries Input:} Uncertainty radius $\delta>0$, parameter $g\in (0,1)$, step-size sequence $\set{\alpha_k:k\in \Z_{\geq 0}}$, termination time $T$ (could be random).
   \STATE {\bfseries Initialization:} $\widehat{Q}_{\delta,0} \equiv 0$, $k=0$.
   \REPEAT
   \FOR{every $(s,a)\in S\times A$}
   
   \STATE Sample independent $N_1,N_2\sim \mathrm{Geo}(g)$. 
   \STATE Independently draw $2^{N_1+1}$ samples $r_i\sim \nu_{s,a}$ and $2^{N_2+1}$ samples $s'_i\sim p_{s,a}$.
   \STATE Compute $\widehat{R}_\delta(s,a)$ and $\widehat{V}_\delta(\widehat{Q}_{\delta,k})(s,a)$ using Equation \eqref{eq.R}-\eqref{def.delta_q}.

  
   
   \ENDFOR
   \STATE Compute $\widehat{\calT}_{\delta,g,k+1}(\widehat{Q}_{\delta,k})=
       \widehat{R}_\delta+\gamma \widehat{V}_\delta(\widehat{Q}_{\delta,k}).$
   \STATE Perform synchronous $Q$-learning update:
    \[
    \widehat{Q}_{\delta,k+1}= (1-\alpha_{t})\widehat{Q}_{\delta,k}+\alpha_{k}\widehat{\calT}_{\delta,g,k+1}(\widehat{Q}_{\delta,k}).\]
     $k\leftarrow k+1$.
   \UNTIL{$k = T$}
\end{algorithmic}
\end{algorithm}
\begin{remark}
The specific algorithm used in \cite{pmlr-v162-liu22a} only has asymptotic guarantees and requires an infinite number of samples to converge, whereas we propose a variant that yields finite-sample guarantees.
More specifically, in the algorithm, we choose $g\in(1/2,3/4)$, while they choose $g\in(0,1/2)$. This has important consequences: each iteration of \cite{pmlr-v162-liu22a}'s algorithm requires an infinite number of samples in expectation, while in our algorithm, the expected number of samples used until iteration $k$ is $n(g)|S||A|k$, where $n(g)$ doesn't depend on $\gamma$ and the MDP instance. 
\end{remark}
\section{Algorithm Complexity}

All proofs to the results in this section are relegated to Sections 2 - 6 in the supplementary materials.

\par Let $(\Omega,\cF,\set{\cF_k}_{k\in\Z_{\geq 0}},\PP)$ be the underlying filtered probability space, where $\cF_{k-1}$ is the $\sigma$-algebra generated by the random variates used before iteration $k$. We motivate our analysis by making the following observations. If we define the noise sequence 
\[
W_{k+1}(\widehat{Q}_{\delta,k}) :=  \widehat{\calT}_{\delta,g,k+1}(\widehat{Q}_{\delta,k})-\calT_{\delta}(\widehat{Q}_{\delta,k}),
\] 
then the update rule of Algorithm \ref{alg.Q_learning} can be written as 
\begin{equation}\label{eq.SA_form_of_DRQ}
\begin{aligned}
\widehat{Q}_{\delta,k+1}=(1-\alpha_k) \widehat{Q}_{\delta,k}+ \alpha_{k}\crbk{\calT_{\delta}(\widehat{Q}_{\delta,k}) + W_{k+1}(\widehat{Q}_{\delta,k})}.
\end{aligned}
\end{equation}
By construction, we expect that under some condition, $\widehat{\calT}_{\delta,g,k+1}(Q)$ is an unbiased estimate of $\calT_{\delta}(Q)$. Hence $\EE[W_{k+1}(\widehat{Q}_{\delta,k})|\cF_k] = 0$. Therefore, Algorithm \ref{alg.Q_learning} has update of the form \eqref{eq.sa_update}, and hence can be analysed as a stochastic approximation. 
\par We proceed to rigorously establish this. First we introduce the following complexity metric parameter:
\begin{definition}\label{def.min_supp_prob}
    Define the \textit{minimal support probability} as
    \small
    \[
    p_\wedge := \inf_{s,a\in S\times A}\sqbk{\inf_{r\in R:\nu_{s,a}(r) > 0}\nu_{s,a}(r)\wedge\inf_{s'\in S:p_{s,a}(s')>0} p_{s,a}(s')}.
    \]
    \normalsize
\end{definition}
The intuition of why the complexity of the MDP should depend on this minimal support probability is that in order to estimate the DR Bellman operator accurately, in worst case one must know the entire support of transition and reward distributions. Therefore, at least $1/ p_\wedge$ samples are necessary. See \cite{si2020distributional} for a detailed discussion. 

\begin{assumption} \label{assump:var_bound_assumptions}
Assume the following holds: 
\begin{enumerate}
\item The uncertainty set size $\delta$ satisfies $\frac{1}{2}p_\wedge\geq 1-e^{-\delta}.$
\item The geometric probability parameter $g\in(0,3/4)$.
\end{enumerate}
\end{assumption}
\begin{remark}The first entry is a technical assumption that ensures the differentiability of the dual form of the robust functional. We use this specific form just for cleanness of presentation. Moreover, we conjecture that such restriction is not necessary to get the same complexity bounds. See the supplement Subsection 6.1 for a detailed discussion. 
\end{remark}
With Assumption \ref{assump:var_bound_assumptions} in place, we are ready to state our key tools and results. The following propositions underly our iteration and expected sample complexity analysis: 
\begin{proposition}\label{prop:Q_infty_norm_bound}
Let $Q_\delta^*$ be the unique fixed point of the DR Bellman operator $\cT_\delta$. Then $\|Q_\delta^*\|_{\infty} \leq r_{\max}({1-\gamma})^{-1}.$
\end{proposition}

\begin{proposition}\label{prop:unbiased} Suppose Assumption \ref{assump:var_bound_assumptions} is in force. For fixed $Q:S\times A\ra \R$, $\widehat \cT_{\delta}(Q)$ is an unbiased estimate of $\cT_{\delta,g}(Q)$; i.e. $\mathbb{E}W(Q) = 0$. 
\end{proposition}
Proposition \ref{prop:unbiased} guarantees the validity of our construction of the unbiased MLMC-DR Bellman estimator. As explained before, this enables us to establish Algorithm \ref{alg.Q_learning} as SA to the fixed point of $\cT_\delta$. 

For simplicity, define the log-order term
\begin{equation}\label{eqn:def_tlide_l}
\begin{aligned}
\tilde l =&    (3+\log (|S||A||R|)\vee \log (|S|^2|A|))^2\\ &\times \log(11/p_\wedge)^2\frac{4(1-g)}{g(3-4g)}.
\end{aligned}
\end{equation}
\begin{proposition}\label{prop:var_sup_bound} Suppose Assumption \ref{assump:var_bound_assumptions} is in force. For fixed $Q$, there exists constant $c > 0$ s.t.
\[
\mathbb{E}\|W(Q)\|_\infty^2\leq\frac{c\tilde l}{\delta^4 p_\wedge^6} \crbk{r_{\max}^2 + \gamma^2\|Q\|_\infty^2}. 
\]
\end{proposition}

Proposition \ref{prop:var_sup_bound} bounds the infinity norm squared of the martingale difference noise. It is central to our  complexity results in Theorems \ref{thm:algo_error_bound} and \ref{thm:sample_complexity}. 
\begin{theorem}\label{thm:algo_error_bound}
Suppose Assumption \ref{assump:var_bound_assumptions} is in force. Running Algorithm \ref{alg.Q_learning} until iteration $k$ and obtain estimator $\widehat{Q}_{\delta,k}$, the following holds:
\par Constant stepsize:
there exists $c,c'> 0$ s.t. if we choose the stepsize sequence
\[
\alpha_k \equiv \alpha\leq \frac{ (1-\gamma)^2\delta^4p_\wedge^6}{c'\gamma^2\tilde l \log (|S||A|)},
\]
then we have
\[
\begin{aligned}
&\EE\|\widehat{Q}_{\delta,k}-Q_\delta^*\|^2_\infty \leq \\
&\frac{3r_{\max}^2}{2(1-\gamma)^2}\crbk{1-\frac{(1-\gamma)\alpha}{2}}^k + \frac{c\alpha r_{\max}^2\log(|S||A|)\tilde l }{\delta^4 p_\wedge^6(1-\gamma)^4 }.
\end{aligned}
\]
\par Rescaled linear stepsize: there exists $c,c'> 0$ s.t. if we choose the stepsize sequence
\[
\alpha_k = \frac{4}{(1-\gamma)(k+K)}, 
\]
where 
\[
K = \frac{c'\tilde l \log(|S||A|)}{\delta^4p_\wedge^6(1-\gamma)^{3}},
\]
then we have
\[
\begin{aligned}
\mathbb{E}\|\widehat{Q}_{\delta,k}-Q_\delta^*\|^2_\infty
\leq  \frac{c r_{\max}^2\tilde l\log(|S||A|)\log(k + K) }{\delta^4 p_\wedge^6(1-\gamma)^5(k+K) }.
\end{aligned}
\]
\end{theorem}
We define the iteration complexity as
\begin{align*}
   k^*(\epsilon) &:= \inf\{k\geq 1 : E\norm{Q_{k} - Q_\delta^*}_\infty^2 \leq \epsilon^2\}.
\end{align*}
The proof of Theorem \ref{thm:algo_error_bound} is based on the recent advances of finite-time analysis of stochastic approximation algorithms \citep{chen2020}. Theorem \ref{thm:algo_error_bound} bounds the algorithmic error by the current iteration completed. This implies an iteration complexity bound, which we will make clear afterwards. 
\par We consider the expected number of samples we requested from the generator to compute the MLMC-DR estimator for one $(s,a)$-pair. It depends on the geometric parameter $g$. Denote this by $n(g)$, then
\[
n(g) =  \mathbb{E}\sqbk{2^{N_1+1}+2^{N_2+1}} =  \frac{4g}{2g-1}.
\]
We define the expected sample complexity $n^*(\epsilon)$ as the total expected number of samples used until $k^*(\epsilon)$ iterations:
$n^*(\epsilon) := |S||A|n(g)k^*(\epsilon).$
Note that when $g > 1/2$, $n(g)$ is finite. This finiteness and Theorem \ref{thm:algo_error_bound} would imply a finite expected sample complexity bound.
\begin{assumption}\label{assump:g_finite_E_nsample}
In addition to Assumption \ref{assump:var_bound_assumptions}, assume $g\in(1/2,3/4)$.
\end{assumption}
\begin{theorem}\label{thm:sample_complexity}
Suppose Assumptions \ref{assump:var_bound_assumptions} and \ref{assump:g_finite_E_nsample} are enforced. The expected sample complexity of Algorithm \ref{alg.Q_learning} for both stepsizes specified in Theorem \ref{thm:algo_error_bound} satisfies
\[
n^*(\epsilon)\lesssim \frac{r_{\max}^2|S||A|}{\delta^4p_\wedge^6(1-\gamma)^5\epsilon^2}.
\]
\end{theorem}
\begin{remark}Theorem \ref{thm:sample_complexity} follows directly from Theorem \ref{thm:algo_error_bound} by choosing the stepsize in the constant stepsize case
\[
\alpha \simeq \frac{ (1-\gamma)^4\delta^4p_\wedge^6\epsilon^2}{\gamma^2\tilde l \log (|S||A|)},
\]
where $\simeq$ and $\lesssim$ mean equal and less or equal up to a log factor and universal constants. The choice of the stepsizes, however, is dependent on $p_{\wedge}$, which is typically unknown a priori. As discussed before, this dependent on $p_{\wedge}$ is an intrinsic source of complexity of the DR-RL problem. So, one direction for future works is to come up with efficient procedure to consistently estimate $p_{\wedge}$. Also, our bound has a $\delta^{-4}$ dependence. However, we believe that it should be $O(1)$ as $\delta\da 0$. Because the algorithmic behavior will converge to that of the classical $Q$-learning
\end{remark}

The sample complexity bound in Theorem \ref{thm:sample_complexity} is not uniform in $g\in (1/2,3/4)$ as we think of $g$ being a design parameter, not an inherit model parameter. The sample complexity dependence on $g$ is $\frac{n(g)4(1-g)}{g(3-4g)}$, minimized at $g \approx 0.64645$. For convenience, we will use $g = 5/8 = 0.625$.

\section{Numerical Experiments}
In this section, we empirically validate our theories using two numerical experiments. Section \ref{sec:hard_MDP} dedicates to hard MDP instances, which are constructed in \cite{li2021q} to prove the lower bound of the standard $Q$-learning. In Section \ref{sec:inventary}, we use the same inventory control problems as the one used in \cite{pmlr-v162-liu22a} to demonstrate the superiority of our algorithms to theirs. More details on the experiment setup are in Section 7 of the supplemental materials.
\subsection{Hard MDPs for $Q$-learning}
\label{sec:hard_MDP}
\begin{figure}[ht]
    \centering
    \includegraphics[width = 0.65\linewidth]{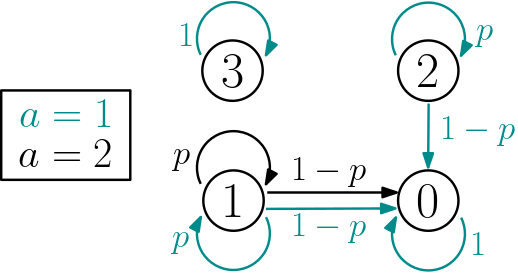}
    \caption{Hard MDP instances transition diagram.}
    \label{fig:hard_mdp_instance}
\end{figure}
First, we will test the convergence of our algorithm on the MDP in Figure \ref{fig:hard_mdp_instance}. It has 4 states and 2 actions, the transition probabilities given action 1 and 2 are labeled on the arrows between states. Constructed in \cite{li2021q}, it is shown that when $p = (4\gamma-1)/(3\gamma)$ the standard non-robust $Q$-learning will have sample complexity $\tilde \Theta((1-\gamma)^{-4}\epsilon^{-2})$. We will use $\delta = 0.1$ in the proceeding experimentation. 
\begin{figure}[ht]\label{fig:hard_conv_test}
\begin{subfigure}{\linewidth}
    \centering
    \includegraphics[width = 0.9\linewidth]{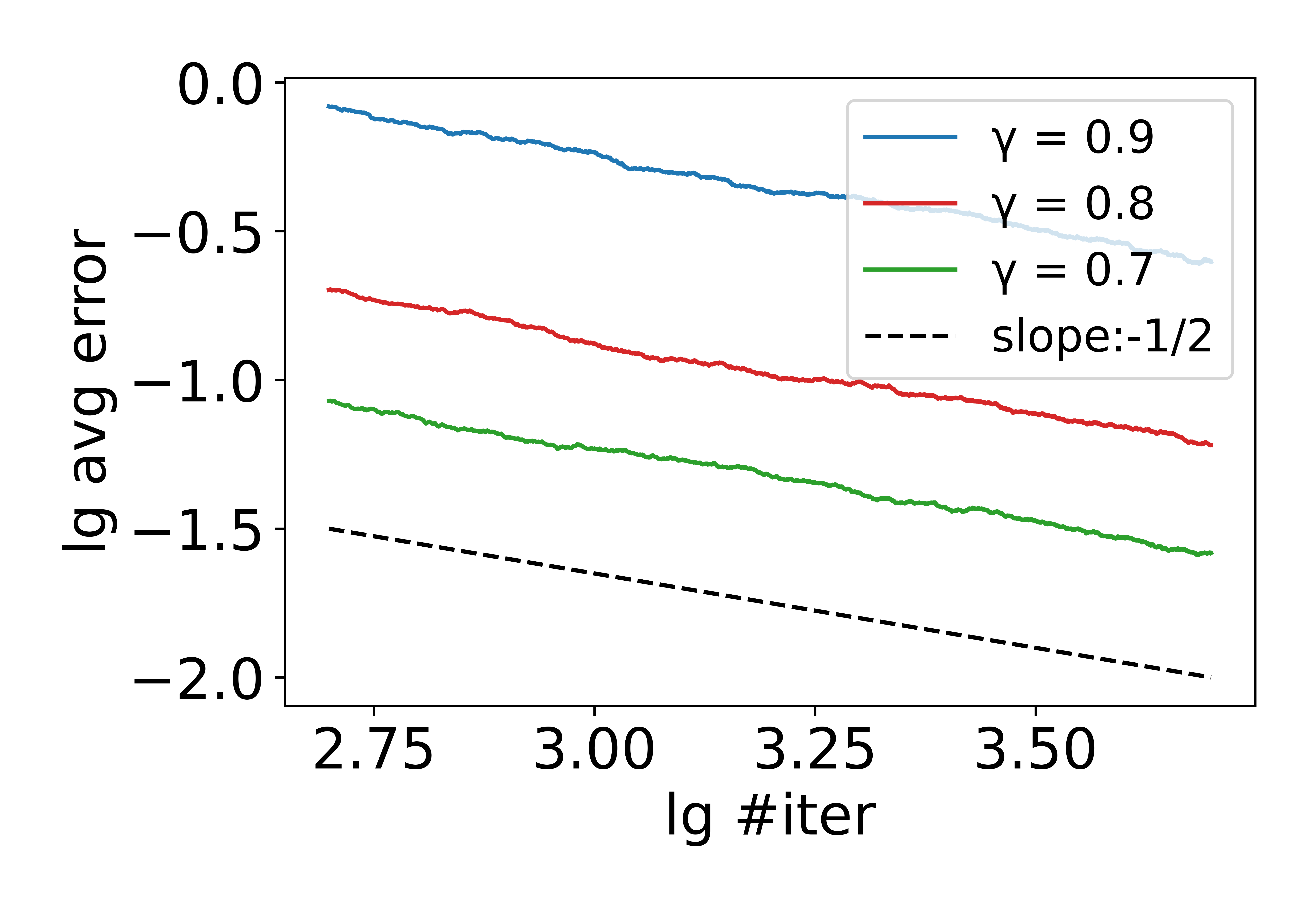}
    \caption{log-log average error with rescaled-linear stepsize.}
    \label{fig:hard_mdp_convergence}
\end{subfigure}
\begin{subfigure}{\linewidth}
    \centering
    \includegraphics[width = 0.9\linewidth]{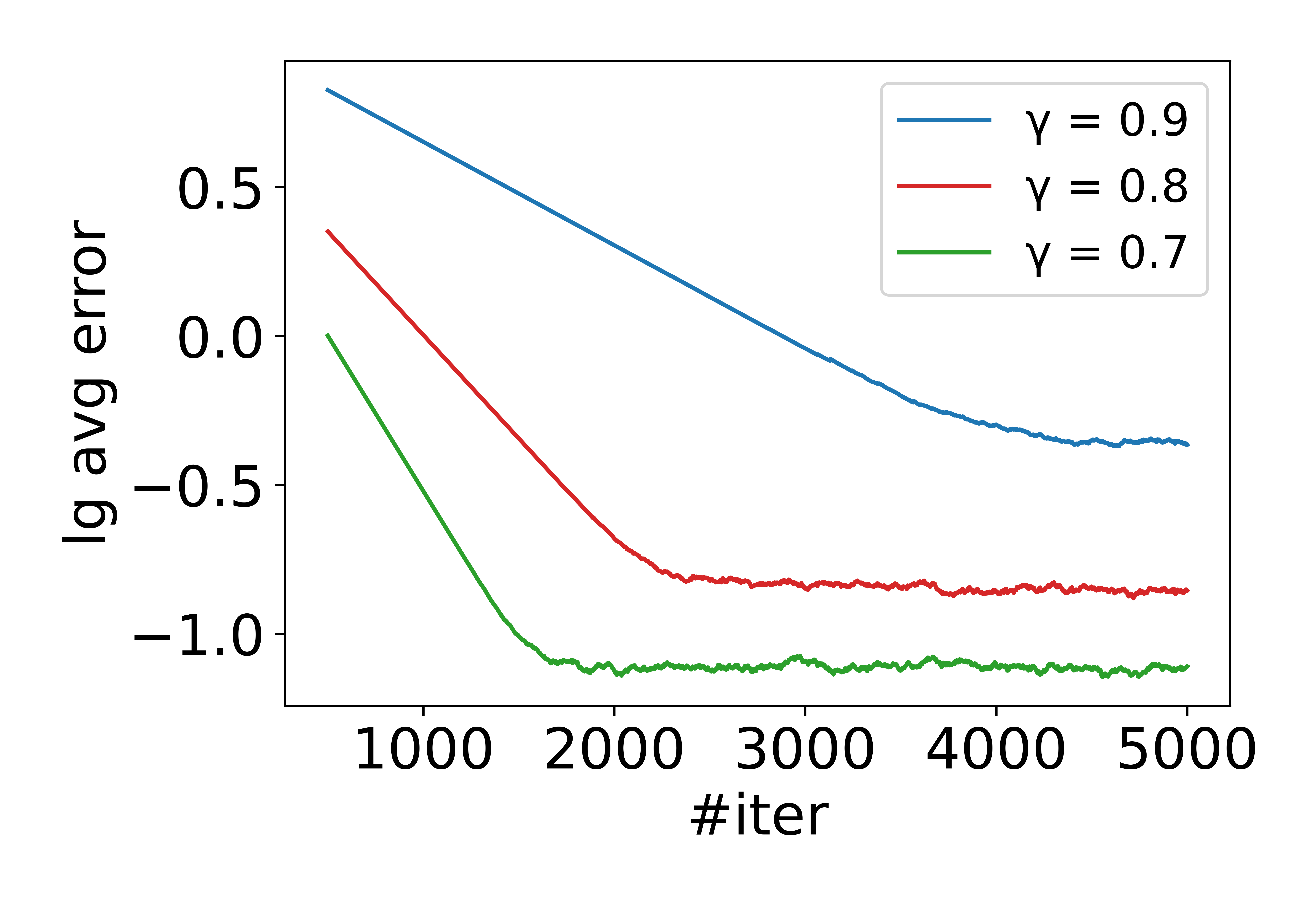}
    \caption{log average error with constant stepsize.}
    \label{fig:hard_mdp_convergence_const}
\end{subfigure}
\caption{Convergence of Algorithm \ref{alg.Q_learning} on MDP \ref{fig:hard_mdp_instance}}
\end{figure}
\par Figures \ref{fig:hard_mdp_convergence} and \ref{fig:hard_mdp_convergence_const} show convergence properties of our algorithm with the rescaled-linear and constant step-size, respectively. Figure \ref{fig:hard_mdp_convergence} is a log-log scale plot of the average (across 200 trajectories) error $\|Q_{\delta,k}-Q^*_\delta\|_\infty$ achieved at a given iteration $k$ under rescaled-linear step-size $\alpha_k = 1/(1+(1-\gamma)k)$. We see that the algorithm is indeed converging. Moreover, Theorem \ref{thm:algo_error_bound} predicts that the slope of each line should be close to $-1/2$, which corresponds to the canonical asymptotic convergence rate $n^{-1/2}$ of the stochastic approximations under the Robbins–Monro step-size regime. This is confirmed in Figure \ref{fig:hard_mdp_convergence}. The algorithm generates Figure \ref{fig:hard_mdp_convergence_const} uses constant step-size $\alpha_k \equiv 0.008$. In Figure \ref{fig:hard_mdp_convergence_const}, the horizontal axis is in linear scale. So, we observe that for all three choices of $\gamma$, the averaged errors first decay geometrically and then stay constant as the number of iterations increases. This is also consistent with the prediction of Theorem \ref{thm:algo_error_bound}.
\begin{figure}[t]
    \centering
    \includegraphics[width =0.9\linewidth]{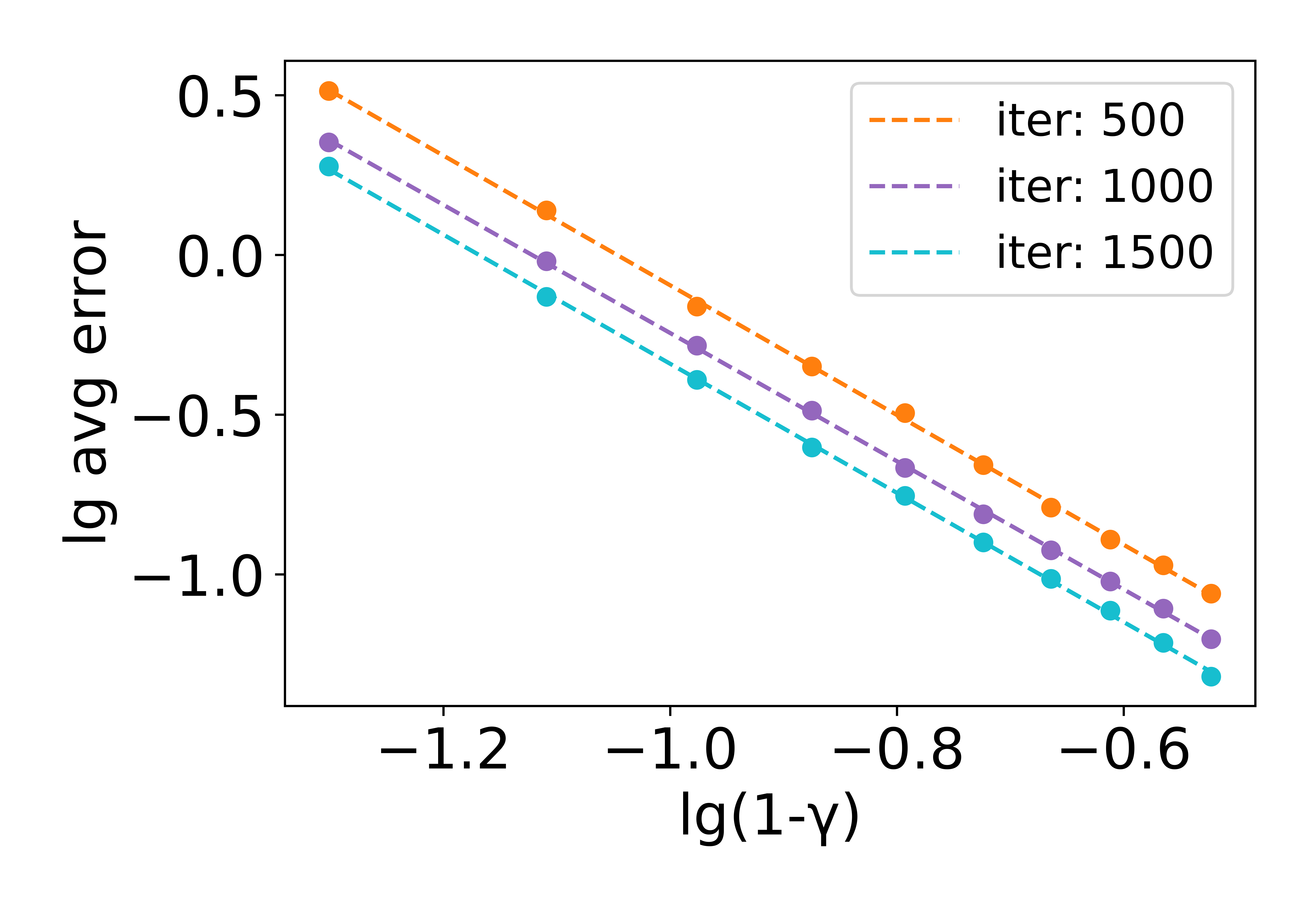}
    \caption{log averaged error against $\log(1-\gamma)$, the slopes of the regression line for iteration $k = 500,1000,1500$ are $-2.031, -2.007, -2.021$.}
    \label{fig:hard_mdp_gamma_dep}
\end{figure}
\par Next, we would like to visualize the $\gamma$-dependence of our algorithm with the rescaled-linear step-size on this hard MDP. Note that if we construct a sequence of hard MDPs with different $(\gamma,Q_{\delta}^*(\gamma),Q^*(\gamma)$, then for fixed iteration $k$, Theorem \ref{thm:algo_error_bound} (ignoring $p_\wedge$ as a function of $\gamma$) implies that $\log E\|Q_{\delta,k}(\gamma)-Q_\delta^*(\gamma)\|_\infty\lesssim - \frac{5}{2}\log (1-\gamma).$
On the other hand the standard $Q$-learning, from \cite{li2021q}, $\log \|Q_{k}(\gamma)-Q^*(\gamma)\|_\infty \asymp - 2\log (1-\gamma)$; corresponding to a $(1-\gamma)^{-4}$ dependence. 
\par Figure \ref{fig:hard_mdp_gamma_dep} plots the average error of the sequence of MLMCDR $Q$-learning at iteration $k = 500, 1000, 1500$ against $\log(1-\gamma)$, and performs a linear regression to extract the slope. We see that for all $k$, the slope is very close to $-2$, suggesting a $(1-\gamma)^{-4}$ dependence. Given that our analysis is based on the finite analysis of SA algorithms in \cite{chen2020}, which, if applied to the classical $Q$-learning, will also yield a $(1-\gamma)^{-5}$ dependence, we think the actual worst case sample complexity is $(1-\gamma)^{-4}$. However, the validity is less clear: the classical non-robust $Q$-learning employs the empirical bellman operator based on one sample each iteration, which is bounded; on the other hand, our estimator uses a random number of samples per iteration, which is only finite w.p.1. This distinction is not visible through the framework of \cite{chen2020} and may result in a rate degradation. 
\par As we point out eariler, we believe that the complexity dependence on $\delta$ should be $O(1)$ as $\delta\da 0$. We also included some experimentation to confirm this in the appendix. 
\subsection{Lost-sale Inventory Control}
\label{sec:inventary}
In this section, we apply Algorithm \ref{alg.Q_learning} to the lost-sale inventory control problem with i.i.d. demand, which is also used in \cite{pmlr-v162-liu22a}.
\par In this model, we consider state and action spaces $S = \set{0,1,\ds,n_s}$, $A = \set{0,1,\ds,n_a}$, $n_a\leq n_s$; the state-action pairs $\set{(s,a)\in S\times A:s + a\leq n_s }$. The demand has support $D = \set{0,1,\ds,n_d}$. We assume that at the beginning of the day $t$, we observe the inventory level $s_t$ and place an order of $a_t$ items which will arrive instantly. The cost incurred on day $t$ is
\begin{align*}
C_t =& k\cd 1\set{A_t > 0}+ h\cd (S_t+A_t - D_t)_+ \\
&+ p\cd (S_t+A_t - D_t)_-
\end{align*}
where $k$ is the ordering cost, $h$ is the holding cost per unit of inventory, and $p$ is the lost-sale price per unit of inventory.  
\par For this numerical experiment, we use $\delta = 0.5$, $\gamma = 0.7$, $n_s = n_a = n_d = 7$,  $k = 3$, $h = 1$, $p = 2$. Under the data-collection environment, we assume $D_t = \unif(D)$ and is i.i.d. across time,
\begin{figure}[t]
\begin{subfigure}{\linewidth}
    \centering
    \includegraphics[width =0.9\linewidth]{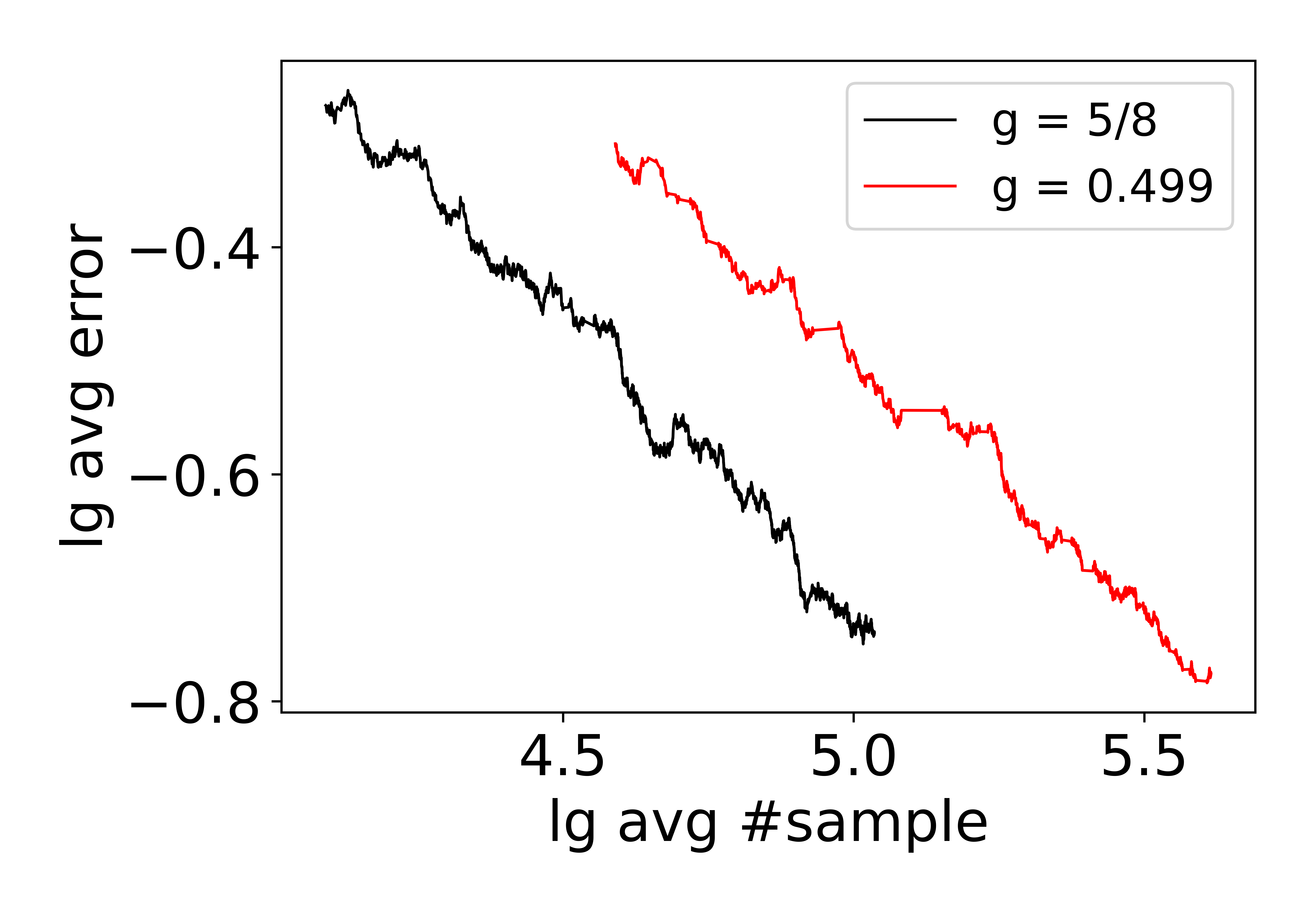}
    \caption{log-log plot of the average error v. the average number of samples at a particular iteration.}
    \label{fig:inventory_avg_avg}
\end{subfigure}
\begin{subfigure}{\linewidth}
    \centering
    \includegraphics[width =0.9\linewidth]{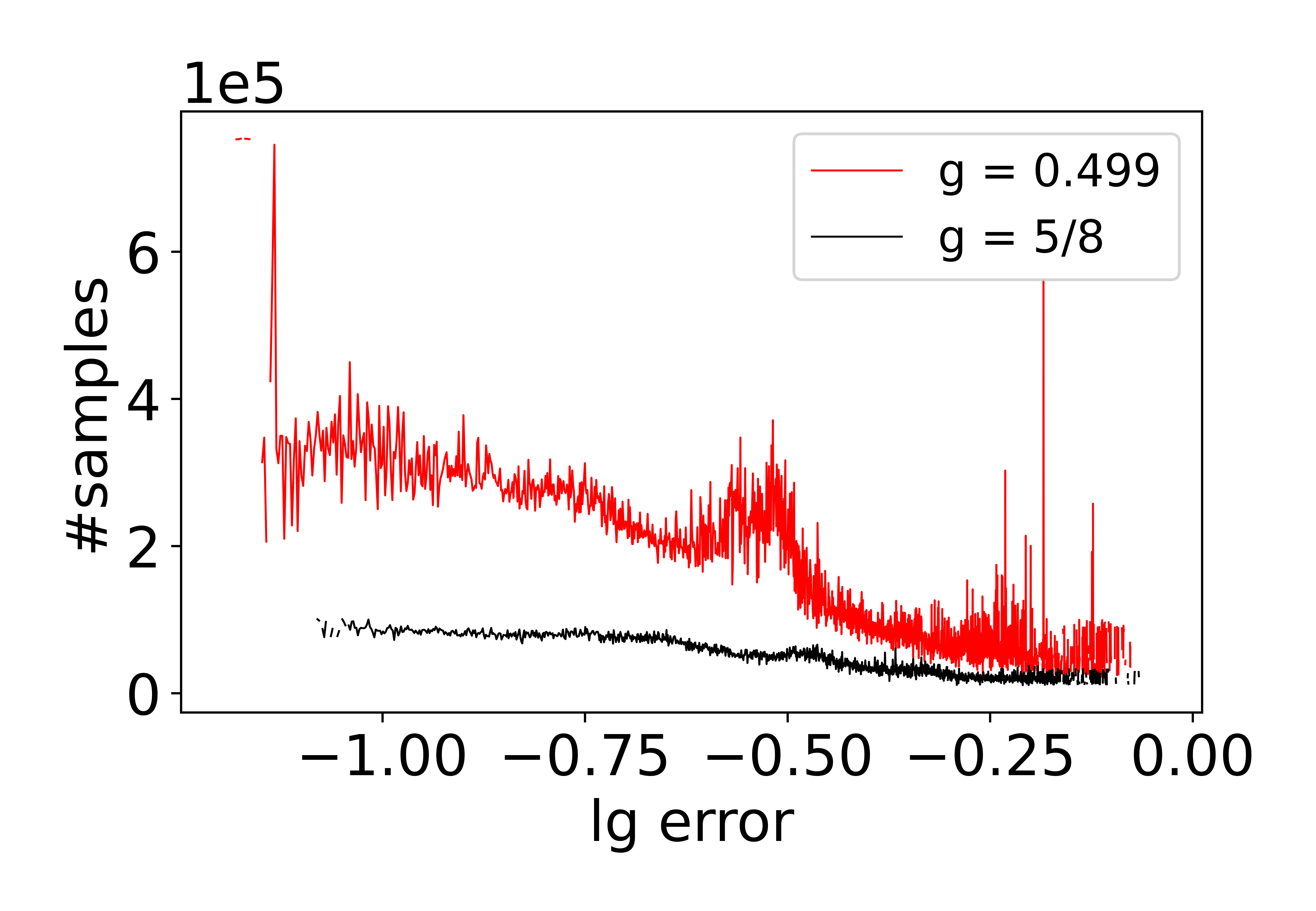}
    \caption{(Smoothed) algorithm error v. number of samples.}
    \label{fig:inventory_sample_at_err}
\end{subfigure}
\caption{Algorithm comparison: inventory model.}
\end{figure}
\par 
Figure \ref{fig:inventory_avg_avg} and \ref{fig:inventory_sample_at_err} compares the sample complexity of  our algorithm to the one proposed in \cite{pmlr-v162-liu22a} ($g\in (1/2,3/4)$ verses. $g\in (0,1/2)$). We run  300 independent trajectories  and record the errors and number of samples used at iteration $1000:5000$.  We clearly observe that our algorithm (black line) outperforms the one in \cite{pmlr-v162-liu22a} (red line).

Specifically, Figure \ref{fig:inventory_avg_avg}, the log-log plot, shows a black line  (our algorithm) of slope close to $-1/2$. This is consistent with our theory and also Figure \ref{fig:hard_mdp_convergence}. The red line (the algorithm in \cite{pmlr-v162-liu22a}) seems to be affine as well with a similar slope. However, there are visible jumps (horizontal segments) along the line. This is due to the MLMC-DR Bellman estimators having infinite mean when $g< 1/2$. 
Furthermore, the overall performance of our algorithm is significantly better. Figure \ref{fig:inventory_sample_at_err} is a (smoothed) scatter plot of the (lg-error, number of sample used) pairs. We can clearly see that not only our algorithm (black line) has better sample complexity performance at every error value, but also the black line has significantly less variation compare to the red line (the algorithm in \cite{pmlr-v162-liu22a}). Again, this is due to our MLMC estimator having a constant order expected sample size in contrast to the infinite expected sample size of the MLMC estimator in \cite{pmlr-v162-liu22a} with the parameter $g < 1/2$. 


\section{Conclusion}
We establish the first model-free finite sample complexity bound for the DR-RL problem:  $\tilde{O}(|S||A|(1-\gamma)^{-5}\epsilon^{-2}\delta^{-4} p_\wedge^{-6})$. Though optimal in $|S||A|$, we believe that the dependence on other parameters are sub-optimal for Algorithm \ref{alg.Q_learning} and need research efforts for further improvements. Also, a minimax complexity lower bound of our algorithm will facilitate a better understanding of its performance. We leave these for future works.  
\acknowledgments{
The work is generously supported by the funding of NSF grants 2312204 and 2312205.

Material in this paper is based upon work supported by the Air Force Office of Scientific Research under award number FA9550-20-1-0397. Additional support is gratefully acknowledged from NSF grants 1915967 and 2118199.

This work is supported in part by National
Science Foundation grant CCF-2106508, and the Digital Twin research grant from Bain \&
Company. Zhengyuan Zhou acknowledges the generous
support from New York University’s 2022-2023 Center for Global Economy and Business
faculty research grant.}

\bibliographystyle{apalike}
\bibliography{DR2,DR4,f_divergence,rates}

\begin{thebibliography}{}

\bibitem[Abadeh et~al., 2018]{abadeh2018Wasserstein}
Abadeh, S.~S., Nguyen, V.~A., Kuhn, D., and Esfahani, P.~M. (2018).
\newblock Wasserstein distributionally robust kalman filtering.
\newblock In {\em Advances in Neural Information Processing Systems}, pages
  8483--8492.

\bibitem[Agarwal et~al., 2020]{pmlr-v125-agarwal20b}
Agarwal, A., Kakade, S., and Yang, L.~F. (2020).
\newblock Model-based reinforcement learning with a generative model is minimax
  optimal.
\newblock In Abernethy, J. and Agarwal, S., editors, {\em Proceedings of Thirty
  Third Conference on Learning Theory}, volume 125 of {\em Proceedings of
  Machine Learning Research}, pages 67--83. PMLR.

\bibitem[Azar et~al., 2013]{azar2013}
Azar, M.~G., Munos, R., and Kappen, H.~J. (2013).
\newblock Minimax pac bounds on the sample complexity of reinforcement learning
  with a generative model.
\newblock {\em Machine learning}, 91(3):325--349.

\bibitem[Bayraksan and Love, 2015]{bayraksan2015data}
Bayraksan, G. and Love, D.~K. (2015).
\newblock Data-driven stochastic programming using phi-divergences.
\newblock In {\em The Operations Research Revolution}, pages 1--19.
  Catonsville: Institute for Operations Research and the Management Sciences.

\bibitem[Bertsekas, 2011]{Bertsekas2012control}
Bertsekas, D.~P. (2011).
\newblock Dynamic programming and optimal control 3rd edition, volume ii.
\newblock {\em Belmont, MA: Athena Scientific}.

\bibitem[Bertsimas and Sim, 2004]{bertsimas2004price}
Bertsimas, D. and Sim, M. (2004).
\newblock The price of robustness.
\newblock {\em Operations Research}, 52(1):35--53.

\bibitem[Blanchet and Murthy, 2019]{blanchet2016quantifying}
Blanchet, J. and Murthy, K. (2019).
\newblock Quantifying distributional model risk via optimal transport.
\newblock {\em Mathematics of Operations Research}, 44(2):565--600.

\bibitem[Blanchet et~al., 2019]{blanchet2019unbiased}
Blanchet, J.~H., Glynn, P.~W., and Pei, Y. (2019).
\newblock Unbiased multilevel monte carlo: Stochastic optimization,
  steady-state simulation, quantiles, and other applications.
\newblock {\em arXiv preprint arXiv:1904.09929}.

\bibitem[Chen et~al., 2018]{wolfram2018}
Chen, Z., Kuhn, D., and Wiesemann, W. (2018).
\newblock Data-driven chance constrained programs over wasserstein balls.
\newblock {\em arXiv preprint arXiv:1809.00210}.

\bibitem[Chen et~al., 2020]{chen2020}
Chen, Z., Maguluri, S.~T., Shakkottai, S., and Shanmugam, K. (2020).
\newblock Finite-sample analysis of contractive stochastic approximation using
  smooth convex envelopes.
\newblock In Larochelle, H., Ranzato, M., Hadsell, R., Balcan, M., and Lin, H.,
  editors, {\em Advances in Neural Information Processing Systems}, volume~33,
  pages 8223--8234. Curran Associates, Inc.

\bibitem[Choi et~al., 2009]{CHOI2009rlSavingBehavior}
Choi, J.~J., Laibson, D., Madrian, B.~C., and Metrick, A. (2009).
\newblock Reinforcement learning and savings behavior.
\newblock {\em The Journal of finance}, 64(6):2515--2534.

\bibitem[Cisneros-Velarde et~al., 2020]{cisneros2020distributionally}
Cisneros-Velarde, P., Petersen, A., and Oh, S.-Y. (2020).
\newblock Distributionally robust formulation and model selection for the
  graphical lasso.
\newblock In {\em International Conference on Artificial Intelligence and
  Statistics}, pages 756--765. PMLR.

\bibitem[Delage and Ye, 2010]{delage2010distributionally}
Delage, E. and Ye, Y. (2010).
\newblock Distributionally robust optimization under moment uncertainty with
  application to data-driven problems.
\newblock {\em Operations research}, 58(3):595--612.

\bibitem[{Deng} et~al., 2017]{Deng2017rlFinanceSignal}
{Deng}, Y., {Bao}, F., {Kong}, Y., {Ren}, Z., and {Dai}, Q. (2017).
\newblock Deep direct reinforcement learning for financial signal
  representation and trading.
\newblock {\em IEEE Transactions on Neural Networks and Learning Systems},
  28(3):653--664.

\bibitem[Drew, 2015]{Abbeel2015DARPA}
Drew, K. (2015).
\newblock California robot teaching itself to walk like a human toddler.
\newblock {\em NBC News}.

\bibitem[Duchi et~al., 2016]{duchi2016statistics}
Duchi, J., Glynn, P., and Namkoong, H. (2016).
\newblock Statistics of robust optimization: A generalized empirical likelihood
  approach.
\newblock {\em arXiv preprint arXiv:1610.03425}.

\bibitem[Duchi et~al., 2019]{duchi2019distributionally}
Duchi, J., Hashimoto, T., and Namkoong, H. (2019).
\newblock Distributionally robust losses against mixture covariate shifts.
\newblock {\em arXiv preprint arXiv:2007.13982}.

\bibitem[Duchi and Namkoong, 2018]{duchi2018learning}
Duchi, J. and Namkoong, H. (2018).
\newblock Learning models with uniform performance via distributionally robust
  optimization.
\newblock {\em arXiv preprint arXiv:1810.08750}.

\bibitem[Even-Dar et~al., 2003]{even2003learning}
Even-Dar, E., Mansour, Y., and Bartlett, P. (2003).
\newblock Learning rates for q-learning.
\newblock {\em Journal of machine learning Research}, 5(1).

\bibitem[Gao and Kleywegt, 2016]{gao2016distributionally}
Gao, R. and Kleywegt, A.~J. (2016).
\newblock Distributionally robust stochastic optimization with {Wasserstein}
  distance.
\newblock {\em arXiv preprint arXiv:1604.02199}.

\bibitem[Gao et~al., 2018]{gao2018robust}
Gao, R., Xie, L., Xie, Y., and Xu, H. (2018).
\newblock Robust hypothesis testing using wasserstein uncertainty sets.
\newblock In {\em Advances in Neural Information Processing Systems}, pages
  7902--7912.

\bibitem[Ghosh and Lam, 2019]{ghosh2019robust}
Ghosh, S. and Lam, H. (2019).
\newblock Robust analysis in stochastic simulation: Computation and performance
  guarantees.
\newblock {\em Operations Research}.

\bibitem[Gonz\'{a}lez-Trejo et~al., 2002]{gonzalez-trejo2002}
Gonz\'{a}lez-Trejo, J.~I., Hern\'{a}ndez-Lerma, O., and Hoyos-Reyes, L.~F.
  (2002).
\newblock Minimax control of discrete-time stochastic systems.
\newblock {\em SIAM Journal on Control and Optimization}, 41(5):1626--1659.

\bibitem[Gu et~al., 2017]{Sergey2017DeepRLinRobotics}
Gu, S., Holly, E., Lillicrap, T., and Levine, S. (2017).
\newblock Deep reinforcement learning for robotic manipulation with
  asynchronous off-policy updates.
\newblock In {\em 2017 IEEE international conference on robotics and automation
  (ICRA)}, pages 3389--3396. IEEE.

\bibitem[Ho-Nguyen et~al., 2020]{ho2020distributionally}
Ho-Nguyen, N., K{\i}l{\i}n{\c{c}}-Karzan, F., K{\"u}{\c{c}}{\"u}kyavuz, S., and
  Lee, D. (2020).
\newblock Distributionally robust chance-constrained programs with right-hand
  side uncertainty under wasserstein ambiguity.
\newblock {\em arXiv preprint arXiv:2003.12685}.

\bibitem[Hu and Hong, 2013a]{Hu2012KullbackLeiblerDC}
Hu, Z. and Hong, L.~J. (2013a).
\newblock Kullback-leibler divergence constrained distributionally robust
  optimization.
\newblock {\em Available at Optimization Online}.

\bibitem[Hu and Hong, 2013b]{hu2013kullback}
Hu, Z. and Hong, L.~J. (2013b).
\newblock Kullback-leibler divergence constrained distributionally robust
  optimization.
\newblock {\em Available at Optimization Online}.

\bibitem[Huang et~al., 2017]{huang2017RLinCV}
Huang, C., Lucey, S., and Ramanan, D. (2017).
\newblock Learning policies for adaptive tracking with deep feature cascades.
\newblock {\em ICCV}, pages 105--114.

\bibitem[Iyengar, 2005]{Iyengar2005}
Iyengar, G. (2005).
\newblock Robust dynamic programming.
\newblock {\em Math. Oper. Res.}, 30:257--280.

\bibitem[Kober et~al., 2013]{Kober2013RLinRoboticsSurvey}
Kober, J., Bagnell, J.~A., and Peters, J. (2013).
\newblock Reinforcement learning in robotics: A survey.
\newblock {\em The International Journal of Robotics Research},
  32(11):1238--1274.

\bibitem[Kushner and Yin, 2013]{kushner2013stochastic}
Kushner, H. and Yin, G. (2013).
\newblock {\em Stochastic Approximation and Recursive Algorithms and
  Applications}.
\newblock Stochastic Modelling and Applied Probability. Springer New York.

\bibitem[Lam, 2019]{lam2019recovering}
Lam, H. (2019).
\newblock Recovering best statistical guarantees via the empirical
  divergence-based distributionally robust optimization.
\newblock {\em Operations Research}, 67(4):1090--1105.

\bibitem[Lam and Zhou, 2017]{lam2017empirical}
Lam, H. and Zhou, E. (2017).
\newblock The empirical likelihood approach to quantifying uncertainty in
  sample average approximation.
\newblock {\em Operations Research Letters}, 45(4):301--307.

\bibitem[Lee and Raginsky, 2018]{raginsky_lee}
Lee, J. and Raginsky, M. (2018).
\newblock Minimax statistical learning with wasserstein distances.
\newblock In {\em Proceedings of the 32nd International Conference on Neural
  Information Processing Systems}, NIPS'18, pages 2692--2701, USA. Curran
  Associates Inc.

\bibitem[Li et~al., 2021]{li2021q}
Li, G., Cai, C., Chen, Y., Gu, Y., Wei, Y., and Chi, Y. (2021).
\newblock Is q-learning minimax optimal? a tight sample complexity analysis.
\newblock {\em arXiv preprint arXiv:2102.06548}.

\bibitem[Li et~al., 2020]{li2020breaking}
Li, G., Wei, Y., Chi, Y., Gu, Y., and Chen, Y. (2020).
\newblock Breaking the sample size barrier in model-based reinforcement
  learning with a generative model.
\newblock In Larochelle, H., Ranzato, M., Hadsell, R., Balcan, M., and Lin, H.,
  editors, {\em Advances in Neural Information Processing Systems}, volume~33,
  pages 12861--12872. Curran Associates, Inc.

\bibitem[Li et~al., 2009]{li2009americanOption}
Li, Y., Szepesvari, C., and Schuurmans, D. (2009).
\newblock Learning exercise policies for american options.
\newblock In {\em Artificial Intelligence and Statistics}, pages 352--359.

\bibitem[Liu et~al., 2022]{pmlr-v162-liu22a}
Liu, Z., Bai, Q., Blanchet, J., Dong, P., Xu, W., Zhou, Z., and Zhou, Z.
  (2022).
\newblock Distributionally robust $q$-learning.
\newblock In Chaudhuri, K., Jegelka, S., Song, L., Szepesvari, C., Niu, G., and
  Sabato, S., editors, {\em Proceedings of the 39th International Conference on
  Machine Learning}, volume 162 of {\em Proceedings of Machine Learning
  Research}, pages 13623--13643. PMLR.

\bibitem[Luenberger et~al., 2021]{luenberger1984linear}
Luenberger, D.~G., Ye, Y., et~al. (2021).
\newblock {\em Linear and nonlinear programming}.
\newblock Springer.

\bibitem[{Maitin-Shepard} et~al., 2010]{Abbeel2010robotFold}
{Maitin-Shepard}, J., {Cusumano-Towner}, M., {Lei}, J., and {Abbeel}, P.
  (2010).
\newblock Cloth grasp point detection based on multiple-view geometric cues
  with application to robotic towel folding.
\newblock In {\em 2010 IEEE International Conference on Robotics and
  Automation}, pages 2308--2315.

\bibitem[Mohajerin~Esfahani and Kuhn, 2018]{MohajerinEsfahani2017}
Mohajerin~Esfahani, P. and Kuhn, D. (2018).
\newblock Data-driven distributionally robust optimization using the
  {Wasserstein} metric: performance guarantees and tractable reformulations.
\newblock {\em Mathematical Programming}, 171(1):115--166.

\bibitem[Namkoong and Duchi, 2016]{namkoong2016stochastic}
Namkoong, H. and Duchi, J.~C. (2016).
\newblock Stochastic gradient methods for distributionally robust optimization
  with f-divergences.
\newblock In {\em Proceedings of the 30th International Conference on Neural
  Information Processing Systems}, pages 2216--2224. Red Hook: Curran
  Associates Inc.

\bibitem[Nguyen et~al., 2018]{nguyen2018distributionally}
Nguyen, V.~A., Kuhn, D., and Esfahani, P.~M. (2018).
\newblock Distributionally robust inverse covariance estimation: The
  {Wasserstein} shrinkage estimator.
\newblock {\em arXiv preprint arXiv:1805.07194}.

\bibitem[Panaganti and Kalathil, 2021]{Panaganti2021}
Panaganti, K. and Kalathil, D. (2021).
\newblock Sample complexity of robust reinforcement learning with a generative
  model.

\bibitem[Powell, 2007]{powell2011approDP}
Powell, W.~B. (2007).
\newblock {\em Approximate Dynamic Programming: Solving the curses of
  dimensionality}, volume 703.
\newblock John Wiley \& Sons.

\bibitem[Sadeghi and Levine, 2016]{Sergey2017RLsingleImage}
Sadeghi, F. and Levine, S. (2016).
\newblock Cad2rl: Real single-image flight without a single real image.
\newblock {\em arXiv preprint arXiv:1611.04201}.

\bibitem[Schulman et~al., 2013]{Abbeel2013robotWalking}
Schulman, J., Ho, J., Lee, A.~X., Awwal, I., Bradlow, H., and Abbeel, P.
  (2013).
\newblock Finding locally optimal, collision-free trajectories with sequential
  convex optimization.
\newblock In {\em Robotics: science and systems}, volume~9, pages 1--10.
  Citeseer.

\bibitem[Shafieezadeh-Abadeh et~al.,
  2015]{shafieezadeh-abadeh_distributionally_2015}
Shafieezadeh-Abadeh, S., Esfahani, P., and Kuhn, D. (2015).
\newblock Distributionally robust logistic regression.
\newblock In {\em Advances in {Neural} {Information} {Processing} {Systems}
  28}, pages 1576--1584.

\bibitem[Shapiro, 2017]{shapiro2017distributionally}
Shapiro, A. (2017).
\newblock Distributionally robust stochastic programming.
\newblock {\em SIAM Journal on Optimization}, 27(4):2258--2275.

\bibitem[Shapiro, 2022]{shapiro2022}
Shapiro, A. (2022).
\newblock Distributionally robust modeling of optimal control.
\newblock {\em Operations Research Letters}, 50(5):561--567.

\bibitem[Shapiro et~al., 2014]{sharpioBookSP}
Shapiro, A., Dentcheva, D., and Ruszczyński, A. (2014).
\newblock {\em Lectures on Stochastic Programming: Modeling and Theory, Second
  Edition}.
\newblock Society for Industrial and Applied Mathematics, Philadelphia, PA.

\bibitem[Shi and Chi, 2022]{Shi2022ModelBase}
Shi, L. and Chi, Y. (2022).
\newblock Distributionally robust model-based offline reinforcement learning
  with near-optimal sample complexity.

\bibitem[Si et~al., 2020]{si2020distributional}
Si, N., Zhang, F., Zhou, Z., and Blanchet, J. (2020).
\newblock Distributional robust batch contextual bandits.

\bibitem[Sidford et~al., 2018]{sidford2018near_opt}
Sidford, A., Wang, M., Wu, X., Yang, L., and Ye, Y. (2018).
\newblock Near-optimal time and sample complexities for solving markov decision
  processes with a generative model.
\newblock In Bengio, S., Wallach, H., Larochelle, H., Grauman, K.,
  Cesa-Bianchi, N., and Garnett, R., editors, {\em Advances in Neural
  Information Processing Systems}, volume~31. Curran Associates, Inc.

\bibitem[Silver et~al., 2016]{alphago2016}
Silver, D., Huang, A., Maddison, C.~J., Guez, A., Sifre, L., Van Den~Driessche,
  G., Schrittwieser, J., Antonoglou, I., Panneershelvam, V., Lanctot, M.,
  et~al. (2016).
\newblock Mastering the game of go with deep neural networks and tree search.
\newblock {\em nature}, 529(7587):484--489.

\bibitem[Silver et~al., 2018]{alphazero2018}
Silver, D., Hubert, T., Schrittwieser, J., Antonoglou, I., Lai, M., Guez, A.,
  Lanctot, M., Sifre, L., Kumaran, D., Graepel, T., et~al. (2018).
\newblock A general reinforcement learning algorithm that masters chess, shogi,
  and go through self-play.
\newblock {\em Science}, 362(6419):1140--1144.

\bibitem[Sinha et~al., 2018]{sinha2018certifiable}
Sinha, A., Namkoong, H., and Duchi, J. (2018).
\newblock Certifiable distributional robustness with principled adversarial
  training.
\newblock In {\em International Conference on Learning Representations}.

\bibitem[Staib and Jegelka, 2017]{staib2017distributionally}
Staib, M. and Jegelka, S. (2017).
\newblock Distributionally robust deep learning as a generalization of
  adversarial training.
\newblock In {\em NIPS workshop on Machine Learning and Computer Security}.

\bibitem[Sutton and Barto, 2018]{sutton2018reinforcement}
Sutton, R.~S. and Barto, A.~G. (2018).
\newblock {\em Reinforcement learning: An introduction}.
\newblock MIT press.

\bibitem[Szepesv{\'a}ri, 2010]{Szepesvari2012rlAlgo}
Szepesv{\'a}ri, C. (2010).
\newblock Algorithms for reinforcement learning.
\newblock {\em Synthesis lectures on artificial intelligence and machine
  learning}, 4(1):1--103.

\bibitem[Volpi et~al., 2018]{volpi2018generalizing}
Volpi, R., Namkoong, H., Sener, O., Duchi, J., Murino, V., and Savarese, S.
  (2018).
\newblock Generalizing to unseen domains via adversarial data augmentation.
\newblock {\em arXiv preprint arXiv:1805.12018}.

\bibitem[Vu et~al., 2022]{vu2022distributionally}
Vu, H., Tran, T., Yue, M.-C., and Nguyen, V.~A. (2022).
\newblock Distributionally robust fair principal components via geodesic
  descents.
\newblock {\em arXiv preprint arXiv:2202.03071}.

\bibitem[Wainwright, 2019a]{wainwright2019l_infty}
Wainwright, M.~J. (2019a).
\newblock Stochastic approximation with cone-contractive operators: Sharp
  $\ell_\infty$-bounds for $q$-learning.

\bibitem[Wainwright, 2019b]{wainwright2019}
Wainwright, M.~J. (2019b).
\newblock Variance-reduced $q$-learning is minimax optimal.

\bibitem[Wiesemann et~al., 2013]{wiesemann2013}
Wiesemann, W., Kuhn, D., and Rustem, B. (2013).
\newblock Robust markov decision processes.
\newblock {\em Mathematics of Operations Research}, 38(1):153--183.

\bibitem[Xu and Mannor, 2010]{huan2010}
Xu, H. and Mannor, S. (2010).
\newblock Distributionally robust markov decision processes.
\newblock In {\em Advances in Neural Information Processing Systems}, pages
  2505--2513.

\bibitem[Yang, 2020]{yang2020Wasserstein}
Yang, I. (2020).
\newblock Wasserstein distributionally robust stochastic control: A data-driven
  approach.
\newblock {\em IEEE Transactions on Automatic Control}.

\bibitem[Yang et~al., 2021]{yang2021}
Yang, W., Zhang, L., and Zhang, Z. (2021).
\newblock Towards theoretical understandings of robust markov decision
  processes: Sample complexity and asymptotics.

\bibitem[Zhao and Guan, 2018]{ZHAO2018262}
Zhao, C. and Guan, Y. (2018).
\newblock Data-driven risk-averse stochastic optimization with {Wasserstein}
  metric.
\newblock {\em Operations Research Letters}, 46(2):262 -- 267.

\bibitem[Zhao and Jiang, 2017]{zhao2017distributionally}
Zhao, C. and Jiang, R. (2017).
\newblock Distributionally robust contingency-constrained unit commitment.
\newblock {\em IEEE Transactions on Power Systems}, 33(1):94--102.

\bibitem[Zhou et~al., 2021]{pmlr-v130-zhou21d}
Zhou, Z., Zhou, Z., Bai, Q., Qiu, L., Blanchet, J., and Glynn, P. (2021).
\newblock Finite-sample regret bound for distributionally robust offline
  tabular reinforcement learning.
\newblock In Banerjee, A. and Fukumizu, K., editors, {\em Proceedings of The
  24th International Conference on Artificial Intelligence and Statistics},
  volume 130 of {\em Proceedings of Machine Learning Research}, pages
  3331--3339. PMLR.

\end{thebibliography}

%

%

\onecolumn
\appendix
\appendixpage
\section{Robust Markov Decision Processes and Reinforcement Learning}
In this section, we rigorously introduce the robust Markov Decision Process formulation and its dynamic programming representation a.k.a. the Bellman equation. Our presentation here is a brief review of the constructions and results in \cite{Iyengar2005}. 
\par We specialize to the infinite horizon, finite state-action-reward MDP setting. In particular recall that $\calM_0 = \left(S, A, R, \calP_0, \calR_0, \gamma\right)$, where $ S$, $ A$, and $R\subsetneq \R_{\geq 0}$ are finite state, action, and reward spaces respectively. $\calP_0= \set{p_{s,a} :s\in S,a\in A}$, $\cR_0 = \set{\nu_{s,a}:{s\in S,a\in A}}$ are the sets of the reward and transition distributions. We consider the KL uncertainty sets $\cR_{s,a}(\delta) = \set{\nu:D_{\text{KL}}(\nu||\nu_{s,a})\leq \delta}$ and $\cP_{s,a}(\delta) = \set{p:D_{\text{KL}}(p||p_{s,a})\leq \delta}$. Note that the robust MDP literature refers to the set $\cP_{s,a}(\delta)$ as ambiguity set so as to distinguish it from randomness (uncertainty). Here, we will follow the convention in DRO literature and use the name uncertainty set. Before moving forward, we define some notations. For discrete set $Y$, let $\cP(Y)$ denote the set of probability measures on $(Y,2^Y)$. Given a probability measure $p\in \cP(Y)$ and a function $f$ on $(Y,2^Y)$, let $p[f]$ denote the integral, i.e. $p[f] = \sum_{y\in Y}p_yf_y$. 
\par We will employ a canonical construction using the product space $\Omega = (S\times A \times R )^{\Z_{\geq 0}}$ and $\cF$ the $\sigma$-field of cylinder sets as the underlying measurable space. At time $t$ the controller at state $s_t$ using action $a_t$ collects a randomized reward $r_t \sim \nu_{s,a}$. 

\par Define the \textit{history} of the controlled Markov chain as the sequence $h_t = (s_0,a_0,s_1,a_1,\ds,s_t)$ and the collection $h_t\in H_t = (S\times A)^{t}\times S$. A history-dependent randomized decision rule at time $t$ is a conditional measure $d_t:H_t\ra \cP(A)$. A decision rule $d_t$ is Markovian if for any $h_t,h'_t\in H_t$ s.t. $h_t = (s_0,a_0,\ds,s_{t-1},a_{t-1},s_t)$ and $h_t = (s_0',a_0',\ds,s_{t-1}',a_{t-1}',s_t)$, we have $d_t(h_t)[\cd] = d_t(h'_t)[\cd]$. A decision rule is deterministic $d_t(h_t)[\1_a] = 1$ for some $a\in A$, where $\1_a$ is the indicator of $\set{a}$, seen as a vector of all 0 but 1 at $a$. The set of transition probabilities consistent with a history dependent decision rule within the uncertainty sets is defined as 
\begin{align*}
\cK^{d_t} = \{ &p:H_t\ra \cP(A\times R\times S): \forall h_t\in H_t,s\in S,a\in A, \\
&p(h_t)[\1_{(a,r,s')}] = d_t(h_t)[\1_a]\nu_{s_t,a}[\1_r]p_{s_t,a}[\1_{s'}] \text{ for some } \nu_{s_t,a}\in \cR_{s_t,a}(\delta),p_{s_t,a}\in \cP_{s_t,a}(\delta)\}.
\end{align*}
Note that this is equivalent to say that $\cR_\delta = \set{\nu_{s,a}:(s,a)\in S\times A}$ is chosen from the product uncertainty set $\prod_{s,a}\cR_{s,a}(\delta)$; same for $p$. This is known as $s,a$-rectangularity; c.f. \cite{wiesemann2013}. Intuitively, $\cK^{d_t}$ is the set of history dependent measures generating a current action $a_t$ from $d_t(h_t)$, and, condition on $s_t,a_t$, independently generate $r_t$ and $s_{t+1}$ from some $\nu\in \cR_{s_t,a}(\delta)$ and $p\in \cP_{s_t,a}(\delta)$. 
\par A history dependent policy $\pi$ is a sequence of decision rules $\pi = (d_t,t\in \Z_{\geq 0})$. We will denote the history-dependent policy class as $\Pi = \set{\pi: (d_t,t\in \Z_{\geq 0})}$. Given a policy $\pi$ we naturally obtain a family of probability measures on $(\Omega,\cF)$; i.e. for initial distribution $\mu\in\cP(S)$
\[
\cK^{\pi}_{\mu}  = \set{P_\mu: \begin{aligned}
 &P(\set{(s_0,a_0,r_0,s_1,a_1,r_1,\ds,r_{T-1},s_T)}) = \mu(s_0)\prod_{t=0}^{T-1}p_k(h_t)[\1_{(a_t,r_t,s_{t+1})}]\\
 &\forall T\in \Z_{\geq 0}, s_i\in S,a_i\in A,r_i\in R, \text{ for some } \set{p_t\in \cK^{d_t}, t\leq T-1}
\end{aligned}
}
\]
where the probabilities of a sample path until time $T$ define the finite dimensional distributions, thence uniquely extend to probability measures on $(\Omega,\cF)$ by the Kolmogorov extension theorem. Formally, we can write $\cK^{\pi} = \prod_{t\geq 0}\cK^{d_t}$. Sometimes, we want to fix the initial state-action $(s_0 = s,a_0 =a)$. This can be done if we let $\mu = \delta_s$ and restrict $\pi$ s.t. $d_0(h_0)[\1_a]$. In order to develop a dynamic programming theory, we assume that the sample path uncertainty set is the full $\cK^{\pi}_\mu$. This is the notion of rectangularity in \cite{Iyengar2005}. 
\par With these constructions, we can rigorously define the \textit{pessimistic} value function $U_\delta^\pi: S\ra \R_+$
\[
U_\delta^\pi(s) := \inf_{P\in \cK_{\delta_s}^\pi}E^P \sqbk{\sum_{t=0}^\infty \gamma^tr_t}
\]
and the optimal pessimistic value function as the following minimax value
\[
U_\delta^*(s) := \sup_{\pi \in \Pi} U_\delta^\pi(s) = \sup_{\pi \in \Pi}\inf_{P\in \cK_{\delta_s}^\pi}E^P \sqbk{\sum_{t=0}^\infty \gamma^tr_t}. 
\]
\par As mentioned before, the rectangularity assumptions allow us to develop a dynamic programming principle, a.k.a. Bellman equation, for the optimal pessimistic value. Before that, we first define and point out some important distinctions to some policy classes. Recall $\Pi$ is the general history dependent policy class. The Markovian deterministic policy class is $\Pi_{\text{MD}} = \set{\pi:d_t \in D_{\text{MD}}}$ where  $D_{\text{MD}}$ is the set of Markovian and deterministic decision rule. The stationary Markovian deterministic policy class $\Pi_{\text{SMD}} = \Pi_{\text{MD}}\cap\set{\pi:d_t = d,\forall t}$. Next, we define the DR Bellman operator for the state value function: 
\begin{align*}
\cB_\delta (v)(s) 
&= \sup_{d_0\in D_{\text{MD}}}\inf_{p\in \cK^{d_0}_{\delta_s}}E_{a,r,s'\sim p}\sqbk{r + \gamma v(s')}\\
&= \sup_{a\in A}\inf_{\nu\in \cR_{s,a}(\delta),p\in \cP_{s,a}(\delta)} E_{r\sim \nu, s'\sim p}[r+\gamma v(s')]
\end{align*}
The second equality follows from noting that the first supremum is achieved by the greedy decision rule. This is our Definition \ref{def.robust_Bellman_opt}; we also recall $V_\delta^*$. Now, we state the dynamic programming principles for the robust MDP: 
\begin{theorem}[Theorem 3.1, 3.2 of \cite{Iyengar2005}]
The following statements hold: 
\begin{enumerate}
    \item $U_\delta^*(s) = \sup_{\pi\in \Pi_{\text{SMD}}} U_{\delta}^\pi(s)$ for all $s\in S$. 
    \item $\cB_\delta$ is a contraction on $(\R^{|S|},\|\cd\|_\infty)$, hence  $V_\delta^*$ is the unique solution to $v = \cB_\delta(v)$. 
    \item $U_\delta^*(s) = V^*_\delta(s)$.
\end{enumerate}
\end{theorem}
\begin{remark}
Our construction allows a randomization of the reward in comparison to \cite{Iyengar2005} and its results easily generalizes. The first entry is a consequence of $S$, $A$ being finite (so that $|\Pi_{\text{SMD}}| < |S|^{|A|}$) and \cite{Iyengar2005}'s Corollary 3.1: for any $\epsilon>0$, $\exists\pi_\epsilon\in\Pi_{\text{SMD}}$ s.t. $\forall s\in S$, $U^*_\delta(s) \leq U_\delta^{\pi_\epsilon}(s) + \epsilon$. 
\end{remark}
This establishes the validity of our approach to the robust control problem by staring from the DR Bellman equation. Of course, the DR $Q$-function, thence the entire DR-RL paradigm, can be interpreted analogous to the classical tabular RL problem. 

\section{Notation and Formulation Remarks}
Let $(\Omega ,\cF,\set{\cF_{k} }, P)$ be a filtered probability space from which we can draw samples, where $\cF_k$ is the smallest $\sigma$-algebra generated by the samples used until iteration $k$ of the algorithm. 

\par Before presenting our proof, we introduce the following notations. For finite discrete measureable space $(Y,2^Y)$, fixed $u\in m2^Y$, and signed measure $m\in \cM_{\pm}(Y,2^Y)$, let $m[u]$ denote the integral. Let $w = w(\alpha) = e^{-u/\alpha}$ and
\begin{equation}\label{eqn:f_def}
f(\mu,u,\alpha) = -\alpha\log \mu[e^{-u/\alpha}]-\alpha\delta.
\end{equation}
We clarify that $f(\mu,u,0) = \lim_{\alpha\da 0}f(\mu,u,\alpha) = \essinf_{\mu}u$. Sometimes, we only need to consider the perturbation analysis on the line of center measures $\set{t\mu_{2^{n}}^O + (1-t)\mu_{2^{n}}^E:t\in[0,1]}$. So, it is convenient to define
\begin{equation}\label{eqn:g,m,mu_def}
\begin{aligned}
\mu_{s,a,n}(t) &= t\mu_{s,a,2^{n}}^O+(1-t)\mu_{s,a,2^{n}}^E\\
m_{s,a,n} &= \mu_{s,a,2^{n}}^O-\mu_{s,a,2^{n}}^E\\
g_{s,a,n}(t,\alpha) &= f(\mu_{s,a,2^{n}}(t),u,\alpha).
\end{aligned}
\end{equation}
Note that we will not explicitly indicate the dependence of $u$ for the function $g$. We will also drop the dependence on $(s,a)$ when clear.
\par Recall that the MLMC DR Bellman operator estimator has the form
\[
\hat \cT_{\delta}(Q)(s,a) = \hat R_{\delta}(s,a) + \gamma \hat V_{\delta}(Q)(s,a)
\]
We want to pursue a unified analysis of $\hat R_{\delta}$ and $\hat V_{\delta}(Q)$. Define the $\Delta$ operator
\begin{align*}
\Delta(\mu_{s,a,2^{n}}^E,\mu_{s,a,2^{n}}^O,u) &= \sup_{\alpha \geq 0} f(\mu_{s,a,2^{n+1}},u,\alpha) - \frac{1}{2}\sup_{\alpha \geq 0} f(\mu_{s,a,2^{n}}^E,u,\alpha)-\frac{1}{2}\sup_{\alpha \geq 0} f(\mu_{s,a,2^{n}}^O,u,\alpha)\\
&= \frac{1}{2}\sqbk{\sup_{\alpha \geq 0}g_{s,a,n}(1/2,\alpha) - \sup_{\alpha \geq 0} g_{s,a,n}(0,\alpha)} +\frac{1}{2}\sqbk{\sup_{\alpha \geq 0}g_{s,a,n}(1/2,\alpha) - \sup_{\alpha \geq 0} g_{s,a,n}(1,\alpha)}.
\end{align*}
Sometimes, we will drop the dependence on $\mu_{s,a,2^{n+1}}$ and $u$ in the following proof. Recall that $Q:S\times A\ra \R$, we define $v(Q): S\ra \R$ by $v(Q)(s) = \max_{a\in A}Q(s,a)$. Let $N_1,N_2\sim p_n = g(1-g)^n$ and $\set{ R_{s,a,i}, S_{s,a,j},i = 0\ds 2^{N_1+1}, j = 0\ds 2^{N_2+1}}$ generated from the reward and transition probabilities; let $\mu^R_{s,a,2^{N_1+1}}$ and $\mu^V_{s,a,2^{N_2+1}}$ be the empirical measures form by the samples $\set{R_{s,a,i},i = 1\ds 2^{N_1+1}}$ and $\set{S_{s,a,i},i = 1\ds 2^{N_2+1}}$ respectively, then under our new notation
\[
\begin{aligned}
\hat R_{\delta}(s,a) &= R_{s,a,0} + \frac{\Delta(\mu^{R,E}_{s,a,2^{N_1}},\mu^{R,O}_{s,a,2^{N_1}},id)}{p_{N_1}}\\
\hat V_{\delta}(Q)(s,a) &= v(Q)(S_{s,a,0}) + \frac{\Delta(\mu^{V,E}_{s,a,2^{N_2}},\mu^{V,O}_{s,a,2^{N_2}},v(Q))}{p_{N_2}}
\end{aligned}
\]
where $id:\R\ra\R$ is the identity function. Note that $\mu^R$ is supported on a finite subset of $\R$, but $\mu^V$ is supported on $S$. This construction suggests that one can employ almost identical analysis on $\hat R_{\delta}$ and $\hat V_{\delta}(Q)$. For notation simplicity, we will write $\Delta_{s,a,n}^R$, $\Delta_{s,a,n}^V$, or the generic $\Delta_{s,a,n}$ when the dependent empirical measures $\mu_{s,a,2^n}^E,\mu_{s,a,2^n}^O$ and function $u$ are contextually clear.
\par Finally, we note that in this notation the minimal support probability definition \ref{def.min_supp_prob} becomes
\[
p_\wedge = \inf_{s,a\in S\times A}\min\set{\inf_{r\in R}\mu_{s,a}^R(r),\inf_{s'\in S} \mu_{s,a}^V(s')}.
\]

\section{Proof of Theorem \ref{thm:algo_error_bound}}
In this section to Theorem \ref{thm:algo_error_bound} assuming the propositions we state ealier. 
\begin{proof}We will denote $E_k[\cd ] = E[\cd|\cF_k]$. Proposition \ref{prop:unbiased} and \ref{prop:var_sup_bound} implies that $E_kW_{k+1}(Q_k) = 0$ and 
\[
E_k\|W_{k+1}(Q_k)\|_\infty^2 \leq \frac{c\tilde l}{\delta^4 p_\wedge^6} \crbk{r_{\max}^2 + \gamma^2\|Q_k\|_\infty^2}. 
\]
Apply Corollary 2.1.1. in \cite{chen2020} under the condition of Corollary 2.1.3, we have that there exists constant $c,c',c'',c'''> 0$ s.t. when 
\[
\alpha_k \equiv \alpha\leq \frac{ (1-\gamma)^2\delta^4p_\wedge^6}{c'\gamma^2\tilde l \log (|S||A|)},
\]
we have
\[
\begin{aligned}
E\|Q_k-Q^*\|^2_\infty &\leq \frac{3}{2}\|Q_0-Q^*\|^2_\infty\crbk{1-\frac{1}{2}(1-\gamma)\alpha}^k + \frac{c''\alpha\log(|S||A|)c\tilde l(r_{\max}^2 + 2\gamma^2 \|Q^*\|_\infty^2)}{\delta^4 p_\wedge^6(1-\gamma)^2 }\\
&\leq \frac{3r_{\max}^2}{2(1-\gamma)^2}\crbk{1-\frac{1}{2}(1-\gamma)\alpha}^k + \frac{c'''\alpha r_{\max}^2\log(|S||A|)\tilde l }{\delta^4 p_\wedge^6(1-\gamma)^4 }.
\end{aligned}
\]
where the last inequality follows from Proposition \ref{prop:Q_infty_norm_bound}. 
Also there exists some other constant $c,c',c'',c'''> 0$ s.t. when 
\[
\begin{aligned}
\alpha_k &= \frac{4}{(1-\gamma)(k+K)}, \\
K &= \frac{c'\tilde l \log(|S||A|)}{\delta^4p_\wedge^6(1-\gamma)^{3}},
\end{aligned}
\]
we have
\[
\begin{aligned}
E\|Q_k-Q^*\|^2_\infty &\leq \frac{3}{2}\|Q_0-Q^*\|^2_\infty \frac{K}{k+K} + \frac{c'' \log(|S||A|)\tilde l (r_{\max}^2 + 2\gamma^2 \|Q^*\|_\infty^2)}{\delta^4 p_\wedge^6(1-\gamma)^3 }\frac{\log(k + K)}{k+K}\\
&\leq  \frac{c''' r_{\max}^2\tilde l\log(|S||A|)\log(k + K) }{\delta^4 p_\wedge^6(1-\gamma)^5(k+K) }. 
\end{aligned}
\]
Renaming the constants gives the theorem. 
\end{proof}

\section{Analysis of the $\Delta$ Operator}
In this section, our analysis uses the compact notation defined in Section 2.
\par Before proving the propositions, we present some key identities of the operator $\Delta_{s,a,n}$. For each $(s,a)\in S\times A$, $\mu_{s,a,2^{n}}^E,\mu_{s,a,2^{n}}^O, \mu_{s,a,2^{n+1}}$ are sampled from $\mu_{s,a}$ the population measure (on finite discrete measureable space $(Y,2^Y)$) and independent across $(s,a)$.  Define for $p > 0$
\[
\Omega_{s,a,n}(p) = \set{\omega: \sup_{y\in Y}|\mu_{s,a,2^{n} }^O(\omega)(y)-\mu_{s,a}(y)|\leq p, \sup_{y\in Y} |\mu_{s,a,2^{n}}^E(\omega)(y)-\mu_{s,a}(y)|\leq p}
\]
We will choose $p = \frac{1}{4}p_\wedge$ where
\[
p_\wedge \leq \inf_{\substack{
(s,a)\in S\times A\\
y:\mu_{s,a}(y) > 0
}} \mu_{s,a}(y)
\]
Note that on $\Omega_{s,a,n}(p)$, for any $(s,a)$
\begin{equation}
\mu_{s,a}\sim \mu_{s,a,2^{n+1}}\sim  \mu_{s,a,2^{n}}^E\sim  \mu_{s,a,2^{n}}^O. \label{eqn:Omega_meas_equiv}
\end{equation} 
Moreover, we have for all $t\in[0,1]$ that could depend on $\omega$,
\begin{equation} \label{eqn:mu_n_dist_on_Omega}
\begin{aligned}
\sup_{y\in Y}|t\mu^E_{s,a,2^n}+(1-t)\mu^O_{s,a,2^n}-\mu|
&\leq t\sup_{y\in Y}|\mu_{s,a,2^n}^E -\mu| + (1-t)\sup_{y\in Y}|\mu_{s,a,2^n}^O -\mu|\\
&\leq p.
\end{aligned}
\end{equation}
\par In this section, we want to bound
\begin{equation}\label{eqn:sup_Delta_2_decomp}
\begin{aligned}
E\sup_{(s,a)\in  S\times A}\Delta_{s,a,n}^2 &= E\sup_{(s,a)\in  S\times A}\Delta_{s,a,n}^2\1_{\Omega_{s,a,n}(p)} + E\sup_{(s,a)\in  S\times A}\Delta_{s,a,n}^2\1_{\Omega_{s,a,n}(p)^c}\\
&=:  E_1+E_2.
\end{aligned}
\end{equation}

\par To bound two terms in equation \eqref{eqn:sup_Delta_2_decomp}, we introduce the following key results:
\begin{lemma}\label{lemma:Delta_n^2_on_Omega_bound} There exists $t\in(0,1)$ s.t.
\[
\Delta_{s,a,n}^2\1_{\Omega_{s,a,n}(p)} \leq \frac{1025\log(11/p_\wedge)^2\umax^2}{\delta^4 p_\wedge^2} \norm{\frac{dm_{s,a,n}}{d\mu_{s,a,n}(t)}}_{L^\infty(\mu)}^4 \1_{\Omega_{s,a,n}(p)}.
\]
\end{lemma}

\begin{lemma}\label{lemma:sub_g_4_max}
Let $\set{Y_i,i=1\ds n}$ be $\sigma^2$ sub-Gaussian, not necessarily independent, then 
\[
EZ:= E\left[\max_{i=1\ds n}Y_i^4\right]\leq 16 \sigma^4 \crbk{3+\log n}^2.
\]
\end{lemma}

\begin{lemma}\label{lemma:sup_f_bound}
For any $ \nu\ll\mu$, we have 
\[
-\|u\|_{L^\infty(\mu)} \leq \sup_{\alpha\geq 0} f(\nu, \alpha) \leq \|u\|_{L^\infty(\mu)}. 
\]
\end{lemma}

\par By \eqref{eqn:mu_n_dist_on_Omega}, 
\[
\inf_{y\in Y}\mu_{s,a,n}(t)(y)\geq p_\wedge - p = \frac{3}{4}p_\wedge. 
\]
So, using \eqref{eqn:sup_Delta_2_decomp} and Lemma \ref{lemma:Delta_n^2_on_Omega_bound}, we can bound
\[
\begin{aligned}
E_1
&\leq \frac{1025\log(11/p_\wedge)^2\umax^2}{\delta^4 p_\wedge^2} E \sup_{(s,a)\in  S\times A} \crbk{\esssup_{\mu} \abs{\frac{dm_{s,a,n}}{d\mu_{s,a,n}(t) }}
}^4\1_{\Omega_{u,n,\epsilon}}\\
&\leq\frac{1025\log(11/p_\wedge)^2\umax^2}{\delta^4 p_\wedge^6} \crbk{\frac{4}{3}}^4  E \sup_{s,a\in S\times A}\sup_{y\in Y} m_{s,a,n}(y)^4
\\
&\leq\frac{3240\log(11/p_\wedge)^2\umax^2}{\delta^4 p_\wedge^6} \frac{1}{2^{4n}}E \sup_{s,a,y} \crbk{\sum_{i=1}^{2^n} \1\set{X_{s,a,i}^O = y}-\1\set{X_{s,a,i}^E =y} }^4\\
\end{aligned}
\]
where $\set{X_{s,a,i}^E,X_{s,a,i}^O,i=1\ds 2^n}$ are independent samples from $\mu_{s,a}$ that forms $\mu_{s,a,n}(t)$. Recall that centered Bernoulli r.v.s. are $1/4$ sub-Gaussian; hence
\[
Y_{s,a}(y):= \sum_{i=1}^{2^n} \1\set{X_{s,a,i}^O = y}-\1\set{X_{s,a,i}^E =y}
\]
is $2^n/2$ sub-Gaussian.  Using lemma \ref{lemma:sub_g_4_max}, we get that there exists constant $c_1$ s.t.
\begin{equation}\label{eqn:E1_bound}
E_1\leq \frac{c_1\log(11/p_\wedge)^2\umax^2}{\delta^4 p_\wedge^6 2^{2n}}  (3+\log (|S||A||Y|))^2
\end{equation}
\par Next, by H\"older's inequality we analyse separately
\[
\begin{aligned}
E_2
&\leq E\left[\sup_{s,a\in S\times A}\Delta_{s,a,n}^2\sup_{s,a\in S\times A}\1_{\Omega_{s,a,n}(p)^c}\right] \\
&\leq \norm{\sup_{s,a\in S\times A}\Delta_{s,a,n}^2}_{L^\infty(P)}P\crbk{\bigcup_{s,a\in S\times A}\Omega_{s,a,n}(p)^c}.
\end{aligned}
\]
Since the empirical measures are sampled from $\mu_{s,a}$, we always have that $\mu_{s,a,n}(t)\ll \mu_{s,a}$. By Lemma \ref{lemma:sup_f_bound}, w.p.1.
\[
\Delta_{s,a,n}^2 \leq 2\|u\|_{L^\infty(\mu_{s,a})} ^2 + 2\|u\|_{L^\infty(\mu_{s,a})} ^2\leq 4\|u\|_{L^\infty(\mu_{s,a})} ^2
\]
where we used Jensen's inequality $(a+b)^2\leq 2a^2 + 2b^2$. Therefore the first term
\[
\norm{\sup_{s,a\in S\times A}\Delta_{s,a,n}^2}_{L^\infty(P)}\leq 4\|u\|_{L^\infty(\mu_{s,a})} ^2. 
\]
For the second term
\[
\begin{aligned}
P\crbk{\bigcup_{s,a\in S\times A}\Omega_{s,a,n}(p)^c}&= P\crbk{\bigcup_{I = E,O}\bigcup_{s,a\in S\times A}\set{\sup_{y\in Y} |\mu_{s,a,2^{n} }^I(y)-\mu_{s,a}(y)|>p}}\\
&\leq 2P\crbk{\bigcup_{s,a\in S\times A}\set{\sup_{y\in Y} |\mu_{s,a,2^{n} }^E(y)-\mu_{s,a}(y)|>p}}\\
&= 2P\crbk{\sup_{s,a\in S\times A}\sup_{y\in Y} |\mu_{s,a,2^{n} }^E(y)-\mu_{s,a}(y)|>p}\\
&\leq \frac{2^9}{p_\wedge^4}E \left[\sup_{s,a,y} |\mu_{s,a,2^{n} }^E(y)-\mu_{s,a}(y)|^4\right]\\
&=\frac{2^9}{p_\wedge^4}\frac{1}{2^{4n}}E\left[ \sup_{s,a,y} \crbk{\sum_{i=1}^{2^n} \1\set{X_{s,a,i}^E = y}- \mu_{s,a}(y) }^4\right]
\end{aligned}
\]
where the second last line follows from Markov's inequality. By Hoeffding's lemma, $\sum_{i = 1}^{2^n}\1\set{X_{s,a,i}^E = y}- \mu_{s,a}(y) $ is $2^n/4$ sub-Gaussian. Therefore, by lemma \ref{lemma:sub_g_4_max}
\[
P\crbk{\bigcup_{s,a\in S\times A}\Omega_{s,a,n}(p)^c} \leq \frac{2^9(3+\log(|S||A||Y|))^2}{p_\wedge^42^{2n}}.
\]
\par Recall \eqref{eqn:sup_Delta_2_decomp}. We conclude that there exists constant $c$ s.t.
\begin{equation} \label{eqn:sup_Delta^2_bound}
\begin{aligned}
E\sup_{(s,a)\in  S\times A}\Delta_{s,a,n}^2 &\leq \frac{c_1\log(11/p_\wedge)^2\umax^2}{\delta^4 p_\wedge^6 2^{2n}}  (3+\log (|S||A||Y|))^2 + \frac{2^{11}(3+\log(|S||A||Y|))^2\umax}{p_\wedge^42^{2n}}\\
&\leq c\log(11/p_\wedge)^2  (3+\log (|S||A||Y|))^2\frac{\umax^2}{\delta^4 p_\wedge^6 2^{2n}}
\end{aligned}
\end{equation}

\section{Proof of Propositions}
With the language and tools developed in the previous sections, we are ready to prove the claimed results
\subsection{Proof of Proposition \ref{prop:contraction}}
\begin{proof}
We use the primal formulation of the distributionally robust Bellman operator
\[
\cT_{\delta}(Q)(s,a) = \inf_{\mu^R_{s,a},\mu^V_{s,a}\sim \delta} \mu^R_{s,a} [id] + \gamma\mu_{s,a}^V\sqbk{v(Q)}.
\]
Then for $Q_1,Q_2\in \R^{S\times A}$,
\[
\begin{aligned}
\cT_{\delta}(Q_1)(s,a) -\cT_{\delta}(Q_2)(s,a)
&= \inf_{\mu^R_{s,a},\mu^V_{s,a}\sim \delta}\crbk{ \mu^R_{s,a} [id] + \gamma\mu_{s,a}^V\sqbk{v(Q_1)}} \\
&\quad + \sup_{\mu^R_{s,a},\mu^V_{s,a}\sim \delta}\crbk{ - \mu^R_{s,a} [id] - \gamma\mu_{s,a}^V\sqbk{v(Q_2)}}\\
&\leq \sup_{\mu^V_{s,a}\sim \delta}\crbk{  \gamma\mu_{s,a}^V\sqbk{v(Q_1)} - \gamma\mu_{s,a}^V\sqbk{v(Q_2)}}\\
&\leq \gamma\sup_{s\in S}(v(Q_1)- v(Q_2)).
\end{aligned}
\]
Therefore,
\[
\begin{aligned}
\|\cT_{\delta}(Q_1) -\cT_{\delta}(Q_2)\|_\infty 
&\leq \gamma\sup_{(s,a)\in S\times A}\max\set{\cT_{\delta}(Q_1)(s,a) -\cT_{\delta}(Q_2)(s,a),\cT_{\delta}(Q_2)(s,a)-\cT_{\delta}(Q_1)(s,a) }\\
&\leq \sup_{(s,a)\in S\times A} \max\set{\sup_{s\in S}(v(Q_1)- v(Q_2)),\sup_{s\in S}(v(Q_2)- v(Q_1))}\\
&= \gamma\sup_{s\in S}|v(Q_1)- v(Q_2)|\\
&= \gamma\sup_{s\in S}\abs{ \sup_{b\in A}Q_1(s,b)- \sup_{b\in A}Q_2(s,b)}\\
&\leq \gamma\sup_{s\in S}\sup_{b\in A}\abs{ Q_1(s,b)- Q_2(s,b)}\\
&\leq \gamma\|Q_1-Q_2\|_\infty;
\end{aligned}
\]
i.e. $\cT_{\delta}$ is a $\gamma$-contraction in $\|\cd\|_\infty$. 
\end{proof}

\subsection{Proof of Proposition \ref{prop:Q_infty_norm_bound}}
\begin{proof}
Recall the dual formulation
\[
\begin{aligned}
\cT_\delta (Q)(s,a)
&=  \sup_{\alpha \geq 0} f(\mu_{s,a}^R,id,\alpha) + \gamma \sup_{\alpha \geq 0} f(\mu_{s,a}^V,v(Q),\alpha)
\end{aligned}
\]
Since $Q^*$ is the fixed point of the distributionally robust Bellman operator. It follows from Lemma \ref{lemma:sup_f_bound} that
\begin{align*}
\|Q^*\|_{\infty} &= \|\cT_\delta(Q^*)\|_{\infty}\\
&\leq \|r\|_\infty + \gamma \|v(Q^*)\|_\infty\\
&\leq r_{\max} + \gamma \|Q^*\|_\infty
\end{align*}
which implies the claimed result. 
\end{proof}
\subsection{Proof of Proposition \ref{prop:unbiased}}
\begin{proof}
By construction, 
\[
\cT_\delta(Q)(s,a) = E\left[R_{s,a,0}\right] + \sum_{n = 0}^\infty E\left[\Delta_{s,a,n}^R\right]+ \gamma E\left[v(Q)(S_{s,a,0})\right] + \gamma \sum_{n = 0}^\infty E\left[\Delta_{s,a,n}^V\right]
\]
If the sum and the integrals can be interchanged, then 
\begin{align*}
\sum_{n = 0}^\infty E\left[\Delta_{s,a,n}^R\right]  &=  E\left[\sum_{n = 0}^\infty\frac{p_n}{p_n}\Delta_{s,a,n}^R\right]\\
&=  E \left[\frac{ \Delta_{s,a,N_1}^R}{p_{N_1}}\right]. 
\end{align*}
Similar results hold for $\hat V_\delta$, and hence the proposition. Therefore, it suffices to exchange the integrals. By Tonelli's theorem, a sufficient condition is that
\[
E\left[\sum_{n = 0}^\infty |\Delta_{s,a,n}^R| \right]< \infty.
\]
Note that by Jensen's inequality,
\begin{align*}
E\left[\sum_{n = 0}^\infty |\Delta_{s,a,n}^R|\right] &= E\left[\sum_{n = 0}^\infty p_n\frac{|\Delta_{s,a,n}^R|}{p_n} \right]\\
&\leq \sqrt{ E\left[\sum_{n = 0}^\infty p_n\frac{{(\Delta_{s,a,n}^R)}
^2}{p_n^2}\right]} \\
&=  \sqrt{ E\left[ \frac{{(\Delta_{s,a,N_1}^R)}
^2}{p_{N_1}^2}\right]}\\
&\leq \sqrt{E\sup_{s,a}\crbk{\frac{{\Delta_{s,a,N_1}^R}
^2}{p_{N_1}^2}}}.
\end{align*}
We show that the quantity in the last line is indeed finite: by Tonelli's theorem, the definition of $\tilde l$ in \eqref{eqn:def_tlide_l}, property \eqref{eqn:sup_Delta^2_bound}, and the choose $g\in(0,3/4)$:
\begin{equation}\label{eqn:E_Delta^2/p^2_r_bound}
\begin{aligned}
E\left[\sup_{s,a} \crbk{\frac{\Delta_{s,a,N_1}^R}{p_{N_1}}}^2\right] &= \sum_{n=0}^\infty \frac{1}{p_{n}}E\sup_{s,a} {\Delta_{s,a,N_1}^R}^2\\
&\leq c\log(11/p_\wedge)^2  (3+\log (|S||A||R|))^2\frac{r_{\max}^2}{\delta^4 p_\wedge^6} \sum_{n=0}^\infty \frac{g}{(4-4g)^n}\\
&= c\log(11/p_\wedge)^2  (3+\log (|S||A||R|))^2\frac{r_{\max}^2}{\delta^4 p_\wedge^6} \frac{4(1-g)}{g(3-4g)}\\
&\leq  \frac{c\tilde l r_{\max}^2}{\delta^4 p_\wedge^6} 
\end{aligned}
\end{equation}
which is finite. Similarly, since $\|v(Q)\|_\infty \leq \|Q\|_\infty$, 
\begin{equation}\label{eqn:E_Delta^2/p^2_v_bound}
E\sup_{s,a} \crbk{\frac{\Delta_{s,a,N_2}^V}{p_{N_2}}}^2 \leq \frac{c\tilde l \|Q\|_\infty^2}{\delta^4 p_\wedge^6}
\end{equation}
is finite as well. This completes the proof
\end{proof}

\subsection{Proof of Proposition \ref{prop:var_sup_bound}}
\begin{proof}Recall bounds \eqref{eqn:E_Delta^2/p^2_r_bound} and \eqref{eqn:E_Delta^2/p^2_v_bound}. We compute
\begin{align*}
E\|W(Q)\|_\infty^2&=E\left[\sup_{s,a} |\hat \cT_{\delta}(Q)(s,a) - \cT_{\delta}(Q)(s,a)|^2\right]\\
&\leq 4 r_{\max}^2 + 4E\sup_{s,a} \crbk{\frac{\Delta_{s,a,N_1}^R}{p_{N_1}}-E \frac{ \Delta_{s,a,N_1}^R}{p_{N_1}}}^2 \\
&\qquad + 8\gamma^2 \|Q\|_\infty^2 + 4\gamma^2 E\sup_{s,a} \crbk{\frac{\Delta_{s,a,N_2}^V}{p_{N_2}}-E \frac{ \Delta_{s,a,N_2}^V}{p_{N_2}}}^2\\
&\leq 4 r_{\max}^2 + 8E\sup_{s,a} \crbk{\frac{\Delta_{s,a,N_1}^R}{p_{N_1}}}^2 +8\sup_{s,a}\crbk{E \frac{ \Delta_{s,a,N_1}^R}{p_{N_1}}}^2 \\
&\qquad + 8\gamma^2 \|Q\|_\infty^2 + 8\gamma^2 E\sup_{s,a} \crbk{\frac{\Delta_{s,a,N_2}^V}{p_{N_2}}}^2+8\gamma^2 \sup_{s,a}\crbk{E \frac{ \Delta_{s,a,N_2}^V}{p_{N_2}}}^2\\
&\leq 4 r_{\max}^2 + 16E\left[\sup_{s,a} \crbk{\frac{\Delta_{s,a,N_1}^R}{p_{N_1}}}^2\right]  + 8\gamma^2 \|Q\|_\infty^2 + 16\gamma^2 E\sup_{s,a} \crbk{\frac{\Delta_{s,a,N_2}^V}{p_{N_2}}}^2\\
&\leq \crbk{4+\frac{16c\tilde l}{\delta^4 p_\wedge^6} }r_{\max}^2 + \crbk{8+\frac{16c\tilde l}{\delta^4 p_\wedge^6} }\gamma^2\|Q\|_\infty^2. 
\end{align*}
Since $\delta < \log 2$, $l > 1$, and $p_\wedge\leq 1/2$, we have
\[
E\|W(Q)\|_\infty^2 \leq \frac{(16c + 4)\tilde l}{\delta^4 p_\wedge^6} r_{\max}^2 + \frac{(16c+8)\tilde l}{\delta^4 p_\wedge^6} \gamma^2\|Q\|_\infty^2. 
\]
replace $16c+8$ with $c$, we obtain the claimed result.
\end{proof}

\section{Proof of Technical Lemmas}
\subsection{Proof of Lemma \ref{lemma:Delta_n^2_on_Omega_bound}}
\begin{proof} Recall the definition \eqref{eqn:f_def} and \eqref{eqn:g,m,mu_def}. We fix $(s,a)$ and write $\Delta_{n} = \Delta_{s,a,n}$, $\mu_{n}(t) = \mu_{s,a,n}(t)$. The following proof assumes that $\omega\in \Omega_{s,a,n}(p)$, so the equivalence \eqref{eqn:Omega_meas_equiv} and the bound \eqref{eqn:mu_n_dist_on_Omega} hold. 
\par From \cite{si2020distributional}, it is sufficient to consider $\alpha\in [0,\delta\inv \|u\|_{L^\infty(\mu)}]=:K$. For $\alpha > 0$ fixed, 
\[
\del_t g_n(t,\alpha) = -\alpha\frac{m_n[w]}{\mu_n(t)[w]}.
\]
Also, for $\alpha = 0$, by \eqref{eqn:Omega_meas_equiv}, $g_n(t,0) \equiv \essinf_\mu u$; hence $\del_t g_n(t,0) \equiv 0$. Let $|m_n|(s) = |m_n(s)|$, again by \eqref{eqn:Omega_meas_equiv}, $\mu_n(t)(s) = 0\iff \mu(s) = 0\implies |m_n|(s) = 0$; i.e. $|m_n|\ll \mu_n(t)$ and the Radon-Nikodym theorem applies. So, for fixed $t\in[0,1]$, 
\begin{align*}
    \lim_{\alpha\da 0} \sup_{s\in (t\pm\epsilon)\cap[0,1]}\abs{\del_t g_n(t,\alpha)} &\leq \lim_{\alpha\da 0}\sup_{t\in [0,1]} \alpha\abs{\frac{m_n[w]}{\mu_n(t)[w]}}\\
&= \lim_{\alpha\da 0} \sup_{t\in [0,1]}\alpha\abs{\frac{1}{\mu_n(t)[w]}\mu_n(t)\sqbk{\frac{dm_n}{d\mu_n(t)}w}}\\
&\leq \lim_{\alpha\da 0} \sup_{t\in [0,1]}\alpha \norm{\frac{dm_n}{d\mu_n(t)}}_{L^\infty(\mu)}\\
&\leq \lim_{\alpha\da 0}  \frac{\alpha}{p_\wedge}\\
&=0.
\end{align*}
where we used H\"older's inequality to get the second last line. Therefore, $\del_tg(\cd,\cd)$ is continuous on $[0,1]\times K$. 
\par Next define
\[
\Theta(t):=\argmax{\alpha\in K}g(t,\alpha). 
\]
We discuss two cases: 
\par \textbf{CASE 1: }If $u$ is $\mu$-essentially constant, then
\[
\sup_{\alpha\in K} -\alpha\log e^{-\bar u/\alpha}-\alpha\delta =\sup_{\alpha\in K} \bar u-\alpha \delta;
\]
i.e. $\Theta(t) = \set{0}$. 
\par \textbf{CASE 2: }$u$ is not $\mu$-essentially constant. Note that when $\alpha > 0$, $w> 0$; we can define a new measure \[
\mu^*_n(t)[\cd] = \frac{\mu_n(t)[w\cd]}{\mu_n(t)[w]}.
\]
We have that
\begin{align*}
\del_\alpha\del_\alpha g_n(t,\alpha) &= - \frac{\mu_n(t)[u^2w] }{\alpha^{3}\mu_n(t)[w]}+\frac{\mu_n(t)[u
w]^2 }{\alpha^{3}\mu_n(t)[w]^2} \\
&= - \frac{\mu^*_n(t)[u^2] }{\alpha^{3}}+\frac{\mu^*_n(t)[u]^2 }{\alpha^3}\\
&= -\frac{\var_{\mu^*_n(t)}(u)}{\alpha^3}\\
&<0; 
\end{align*}
i.e. $g_n(t,\cd)$ is strictly concave for $\alpha > 0$. Also, recall that $g_n(t,\cd)$ is continuous at $0$. So, in this case either $\Theta(t) =\set{0}$ or $\Theta(t) = \set{\alpha_n^*(t)}$ where $\delta\inv \|u\|_{L^\infty(\mu)}\geq \alpha_n^*(t) > 0 $. 
\par In particular, $\Theta(t)$ is a singleton which we will denote by $\alpha_n^*(t)$ in both cases. We conclude that by \cite{sharpioBookSP} Theorem 7.21, the following derivative exists
\[
d_t\sup_{\alpha\in K}g_n(t,\alpha) = \sup_{\alpha\in \Theta(t)}\del_t g_n(t,\alpha) = \del_t g_n(t,\alpha_n^*(t)).
\]
Therefore, by the mean value theorem, there exists $t_1\in (0,1/2),t_2\in(1/2,1)$ depending on $\omega$ s.t. 
\begin{align*}
\Delta_n &= \frac{1}{2}\crbk{ \del_t g_n(t_1,\alpha_n^*(t_1)) - \del_t g_n(t_2,\alpha_n^*(t_2))}\\
&= -\frac{1}{2}\crbk{ \alpha^*_n(t_1)\frac{m_n[w^*_n(t_1)]}{\mu_n(t_1)[w^*_n(t_1)]} - \alpha^*_n(t_2)\frac{m_n[w^*_n(t_2)]}{\mu_n(t_2)[w^*_n(t_2)]}}.
\end{align*}
where $w_n^*(t) = e^{-u/\alpha_n^*(t)}$. We will use $w$ to denote $w^*_n(t)$ in the following derivations.
\par Again if $u$ is $\mu$-essentially constant, then $\Delta_n = 0$. If not, then we consider the population optimizer, which, by the same reasoning, is also a singleton denoted by $\alpha^*$. Let $\kappa_{s,a} = \mu_{s,a}(\set{s :u(s) = \essinf_{\mu_{s,a}} u})$. There are two cases
\begin{enumerate}
    \item $\alpha^* = 0$. From \cite{hu2013kullback}, $\alpha^* = 0$ iff $ \kappa_{s,a} \geq e^{-\delta}$. If we want $\alpha^*_n(t) = 0$ for all $t\in[0,1]$, a sufficient condition is that $\kappa_{s,a,n}(t) \geq \kappa_{s,a} - p\geq e^{-\delta}$. 
    \item $\alpha^* >0$ iff $ \kappa_{s,a} < e^{-\delta}$. If we want $\alpha^*_n(t) > 0$ for all $t\in[0,1]$, a sufficient condition is that $\kappa_{s,a,n}(t)\leq \kappa_{s,a} + p< e^{-\delta}$. 
\end{enumerate}
Therefore, for any $e^{-\delta}\neq \set{\kappa_{s,a}:(s,a)\in S\times A} \subset\set{\mu_{s,a}(y):(s,a,y)\in S\times A \times Y}$. We can always choose $p$ small enough s.t. for $\omega\in\Omega_{s,a,n}(p)$, $\alpha^* = 0$ or $\alpha^* > 0$ implies that $\alpha_n^*(t) = 0$ or $\alpha_n^*(t) > 0$ respectively. 

\begin{remark}
While this generalizes to all but finitely many $\delta$, for simplicity of presentation, we assume Assumption \ref{assump:var_bound_assumptions} that $p_\wedge/2 \geq 1-e^{-\delta}$. This implies that if $\kappa_{s,a}\neq 1$, then $1-\kappa_{s,a} \geq p_{\wedge}> 1-e^{-\delta}$; i.e. $\kappa_{s,a} < e^{-\delta}$ and case 1 cannot happen. Moreover, if we choose $p = \frac{1}{4}p_\wedge$, then 
\[
\kappa_{s,a}+p\leq 1-\frac{3}{4}p_\wedge < 1-\frac{1}{2}p_\wedge\leq e^{-\delta}
\]
satisfying the sufficient condition in case 2.
\end{remark} Therefore, for $\omega\in\Omega_{s,a,n}(p)$, there are two cases: 
\par \textbf{CASE 1: }$\alpha^* = 0$, then $\Delta_n = 0$, Lemma \ref{lemma:Delta_n^2_on_Omega_bound} holds trivially. 
\par \textbf{CASE 2: } $\alpha^* >0$, then $\alpha^*_n(t_1), \alpha^*_n(t_2)> 0$. Since $g_n(t,\cd)$ is strictly convex, $\alpha^*_n(t)$ is the unique solution to the first order optimality condition
    \begin{equation}
    0 = \del_\alpha g_n(t,\alpha_n^*(t)) = -\log\mu_n(t)[w] -\delta - \frac{\mu_n(t)[uw]}{\alpha_n^*(t)\mu_n(t)[w]}. \label{eqn:alpha_opt_cond}
    \end{equation}
    Note that $\del_\alpha g_n\in C^\infty([0,1]\times \R_{++})$ and that $\del_\alpha\del_\alpha g_n(t,\alpha_n^*(t)) <0$. The implicit function theorem  implies that $\alpha_n^*(t)\in C^1((0,1))$ with derivative
    \begin{align*}
    d_t\alpha_n^*(t) &= -\frac{\del_t\del_\alpha g_n(t,\alpha_n^*(t))}{\del_\alpha\del_\alpha g_n(t,\alpha_n^*(t))}\\
    &=\crbk{\frac{\alpha_n^*(t)^3}{\var_{\mu^*_n(t)}(u)}}\crbk{-\frac{m_n[w]}{\mu_n(t)[w]}  + \frac{\mu_n(t)[uw] m_n[w]}{\alpha_n^*(t)\mu_n(t)[w]^2} - \frac{m_n[uw]}{\alpha_n^*(t)\mu_n(t)[w]}}\\
    &=\crbk{\frac{\alpha_n^*(t)^3}{\var_{\mu^*_n(t)}(u)}}\crbk{-\frac{m_n[w]}{\mu_n(t)[w]}  + \frac{\mu_n(t)[uw/\alpha_n^*(t)] }{\mu_n(t)[w]^2}\mu_n(t)\sqbk{\frac{dm_n}{d\mu_n(t)} w} - \frac{\mu_n(t)\sqbk{\frac{dm_n}{d\mu_n(t)} uw/\alpha_n^*(t)}}{\mu_n(t)[w]}}\\
    &=\crbk{\frac{\alpha_n^*(t)^3}{\var_{\mu^*_n(t)}(u)}}\crbk{- \frac{m_n[w]}{\mu_n(t)[w]} + \mu^*_n(t)[u/\alpha_n^*(t)] \mu^*_n(t)\sqbk{\frac{dm_n}{d\mu_n(t)}} - \mu^*_n(t)\sqbk{\frac{dm_n}{d\mu_n(t)} u/\alpha_n^*(t)}}\\
    &=-\crbk{\frac{\alpha_n^*(t)^3}{\var_{\mu^*_n(t)}(u)}}\crbk{ \frac{m_n[w]}{\mu_n(t)[w]} +\cov_{\mu^*_n(t)}\crbk{\frac{u}{\alpha_n^*(t)},\frac{dm_n}{d\mu_n(t)}}}
    \end{align*} Therefore, we conclude that 
    \[ 
    \del_t g_n(t,\alpha_n^*(t)) = -\alpha^*_n(t)\frac{m_n[w]}{\mu_n(t)[w]}
    \]
    is $C^1((0,1))$ as a function of $t$ with derivative 
    \begin{align*}
    -d_t\del_t g_n(t,\alpha_n^*(t)) &= \alpha_n^*(t)\frac{m_{n}[w]^2}{\mu_{n}(t)[w]^2} - d_t\alpha_n^*(t)\crbk{\frac{m_n[w]}{\mu_n(t)[w]}+ \frac{m_n[uw]}{\alpha_n^*(t)\mu_n(t)[w]}- \frac{m_n[w]\mu_n(t)[uw]}{\alpha_n^*(t)\mu_n(t)[w]^2}}\\
    &= \alpha_n^*(t)\frac{m_{n}[w]^2}{\mu_{n}(t)[w]^2}  + \crbk{\frac{\alpha_n^*(t)}{\var_{\mu^*_n(t)}(u/\alpha_n^*(t))}}\crbk{\mu^*_n(t)\sqbk{\frac{dm_n}{d\mu_n(t)}}+ \cov_{\mu^*_n(t)}\crbk{\frac{u}{\alpha_n^*(t)},\frac{dm_n}{d\mu_n(t)}}}^2.
    \end{align*}
    Therefore, by the mean value theorem, there exists $t_3\in(t_1,t_2)$
    \begin{equation}\label{eqn:|Delta_n|_bound_on_Omega}
    \begin{aligned}
    |\Delta_n| &= \frac{|t_1-t_2|}{2}\abs{d_t\del_t g_n(t,\alpha_n^*(t))|_{t_3}}\\
    &\leq \crbk{\alpha_n^*(t)+ \frac{\alpha_n^*(t)}{\var_{\mu^*_n(t)}(u/\alpha_n^*(t))}}\frac{m_{n}[w]^2}{\mu_{n}(t)[w]^2} + \alpha_n^*(t)\var_{\mu^*_n(t)}\crbk{\frac{dm_n}{d\mu_n(t)}}\Bigg|_{t_3}\\
    &= \crbk{\alpha_n^*(t)+ \frac{\alpha_n^*(t)}{\var_{\mu^*_n(t)}(u/\alpha_n^*(t))}}\frac{m_{n}[w]^2}{\mu_{n}(t)[w]^2} \\
    &\quad + \alpha_n^*(t)\frac{\mu_n(t)\sqbk{\crbk{\frac{dm_n}{d\mu_n(t)}}^2w}}{\mu_n(t)[w]} - \alpha_n^*(t)\frac{\mu_n(t)\sqbk{\frac{dm_n}{d\mu_n(t)}w}^2}{\mu_n(t)[w]^2}\Bigg|_{t_3}\\
    &=  \frac{\alpha_n^*(t_3)}{\var_{\mu^*_n(t_3)}(u/\alpha_n^*(t_3))}\frac{m_{n}[w]^2}{\mu_{n}(t_3)[w]^2} + \alpha_n^*(t_3)\mu^*_n(t_3)\sqbk{\crbk{\frac{dm_n}{d\mu_n(t_3)}}^2}
    \end{aligned}
    \end{equation}
Note that the log-likelihood ratio
\[
\log \crbk{\frac{w}{\mu_n(t)[w]} }= -\frac{u}{\alpha_n^*(t) } -\log \mu_n(t)[w]. 
\]
Moreover, by the optimality condition \eqref{eqn:alpha_opt_cond}, 
\begin{align*}
    \mu^*_n(t)\sqbk{\log \crbk{\frac{w}{\mu_n(t)[w]} }} 
    &= \mu^*_n(t)\sqbk{-\frac{u}{\alpha_n^*(t)}-\log {\mu_n(t)[w] }}\\
    &= \mu^*_n(t)\sqbk{-\frac{u}{\alpha_n^*(t)} }+ \delta + \frac{\mu_n(t)[uw]}{\alpha_n^*(t)\mu_n(t)[w]}\\
    &=- \mu^*_n(t)\sqbk{\frac{u}{\alpha_n^*(t)}} + \delta + \mu^*_n(t)\sqbk{\frac{u}{\alpha_n^*(t)}}\\
    &= \delta.
\end{align*}
So, the variance: 
\begin{align*}
\var_{\mu^*_n(t)}(u/\alpha_n^*(t)) &= \var_{\mu^*_n(t)}\crbk{\log \crbk{\frac{w}{\mu_n(t)[w]} }}\\
&= \mu^*_n(t)\sqbk{\log \crbk{\frac{d\mu^*_n(t)}{d\mu_{n}(t)}}^2} - \delta^2.
\end{align*}
We bound this expression by the following lemma: 
\begin{lemma}\label{lemma:var_lower_bound}
For measures $\mu$, $\mu'$ s.t. $D_{\text{KL}}(\mu'||\mu) = \delta$ and $\bar p_\wedge = \inf_{s\in S}\mu(s) = 1-e^{-\delta-\psi_\wedge}$ for some $\psi_\wedge > 0$ we have that
\[
\mu'\sqbk{\log \crbk{\frac{d\mu'}{d\mu}}^2} - \delta^2 \geq -\frac{\delta^2\psi_\wedge}{8\log(\psi_\wedge/8)}
\]
\end{lemma}

Recall that we choose $p = \frac{1}{4}p_\wedge$, so we should choose $\psi_\wedge$
\[
\bar p_\wedge = 1-e^{-\delta - \psi_\wedge} = \inf_{s:\mu(s) > 0}\mu_n(t)(s) \geq \frac{3}{4}p_\wedge.
\]
We want the above bound to hold uniformly in $\delta$:
\[
\frac{3}{4}p_\wedge \leq \inf_{\delta\geq 0}1-e^{-\delta - \psi_\wedge} = 1-e^{ - \psi_\wedge} 
\]
So, 
\[
\psi_\wedge \geq -\log\crbk{1-\frac{3}{4}p_\wedge}\geq \frac{3}{4}p
_\wedge. 
\]
We conclude that by Lemma \ref{lemma:var_lower_bound}, 
\begin{align*}
\var_{\mu^*_n(t)}(u/\alpha_n^*(t)) 
&= \mu^*_n(t)\sqbk{\log \crbk{\frac{d\mu^*_n(t)}{d\mu_{n}(t)}}^2} - \delta^2\\
&\geq -\frac{3}{32}\frac{\delta^2 p_\wedge}{\log(3 p_\wedge /32)}
\end{align*}
Note that $-x/\log x = O(x^{1+\epsilon})$ as $x\da 0$ for any $\epsilon > 0$. 
\par Next we go back to bounding $\Delta_n$ in case 2. 
\begin{lemma}\label{lemma:alpha_ratio_bound}
For $\delta\leq \log 2/2$ and $\omega\in\Omega_{s,a,n}(\frac{1}{4}p_\wedge)$, we have
\begin{align*}
\sup_{\alpha\in K}\frac{\alpha m_{n}[w]^2}{\mu_{n}(t_3)[w]^2}
\leq  3\umax \norm{\frac{dm_n}{d\mu_n(t_3)}}_{L^\infty(\mu)}^2.
\end{align*}
\end{lemma}

\par We conclude that from \eqref{eqn:|Delta_n|_bound_on_Omega} and Lemma \ref{lemma:alpha_ratio_bound}
\[
\begin{aligned}
\Delta_n^2 \1_{\Omega_{s,a,n}(p)}
&\leq \crbk{\frac{2^{10}\log(\frac{32}{3 p_\wedge} )^2}{9\delta^4 p_\wedge^2}\frac{\alpha_n^*(t_3)^2m_{n}[w]^4}{\mu_{n}(t_3)[w]^4} +  \alpha_n^*(t_3)^2\mu^*_n(t_3)\sqbk{\crbk{\frac{dm_n}{d\mu_n(t_3)}}^2}^2} \1_{\Omega_{s,a,n}(p)}\\
&\leq \crbk{\frac{2^{10}\log(11/p_\wedge)^2}{9\delta^4 p_\wedge^2}\crbk{\sup_{\alpha\in K}\frac{\alpha m_{n}[w]^2}{\mu_{n}(t_3)[w]^2}}^2 +  \crbk{\sup_{\alpha\in K}\alpha\mu^*_n(t_3)\sqbk{\crbk{\frac{dm_n}{d\mu_n(t_3)}}^2}}^2} \1_{\Omega_{s,a,n}(p)}\\
&\leq \crbk{\frac{2^{10}\log(11/p_\wedge)^2}{\delta^4 p_\wedge^2}\umax^2 \norm{\frac{dm_n}{d\mu_n(t_3)}}_{L^\infty(\mu)}^4 +  \frac{\|u\|_{\infty}^2 }{\delta^2}\norm{\frac{dm_n}{d\mu_n(t_3)}}_{L^\infty(\mu)}^4} \1_{\Omega_{s,a,n}(p)}\\
&\leq \frac{1025\log(11/p_\wedge)^2}{\delta^4 p_\wedge^2}\umax^2 \norm{\frac{dm_n}{d\mu_n(t_3)}}_{L^\infty(\mu)}^4 \1_{\Omega_{s,a,n}(p)}.\\
\end{aligned}
\]
This completes the proof of Lemma \ref{lemma:Delta_n^2_on_Omega_bound}.
\end{proof}
\subsubsection{Proof of Lemma \ref{lemma:var_lower_bound}}
\begin{proof}
Consider the program for the probability measure $\nu$ defined on the positive entries of $\mu$, i.e., $\nu = \set{\nu(s)\geq 0 \text{ for } \mu(s) > 0,s\in S; \nu(s)= 0 \text{ for } \mu(s) = 0,s\in S; \sum_{s\in S} \nu(s) = 1}$:
\[
OPT_1 = \inf_{\nu:D_{\text{KL}}(\nu||\mu) =\delta} \nu\sqbk{\log \crbk{\frac{d\nu}{d\mu}}^2}.
\]
We first show that any feasible $\nu$ must satisfy $\mu \sim \nu$; i.e. if $\mu(s) > 0$ then $\nu(s) > 0$. Suppose, to the contrary that there exists $A\in\cS$ s.t. $\mu(A^c) > 0$ but $\nu(A^c) = 0$. Also, since $\mu\gg\nu$, $\nu(A) = 1$ implies that $\mu(A) > 0$. We can define $\mu_A(\cd) = \mu(A\cap \cd)/\mu(A)$ the conditional measure.
\begin{align*}
\delta &= D_{\text{KL}}(\nu||\mu)\\
&= \nu\sqbk{\log\crbk{\frac{d\nu}{d\mu}}}\\
&= \mu \sqbk{\1_A\frac{d\nu}{d\mu}\log\crbk{\frac{d\nu}{d\mu}}}\\
&= \mu(A)\mu_A \sqbk{\frac{d\nu}{d\mu}\log\crbk{\frac{d\nu}{d\mu}}}
\end{align*}
Note that the function $x\ra x\log x$ is convex for $x \geq 0$. We have that by Jensen's inequality
\begin{align*}
\delta &\geq \mu(A)\mu_A \sqbk{\frac{d\nu}{d\mu}}\log \crbk{\mu_A \sqbk{\frac{d\nu}{d\mu}}}\\
&= \mu \sqbk{\1_A\frac{d\nu}{d\mu}}\log \crbk{\frac{1}{\mu(A)}\mu \sqbk{\1_A\frac{d\nu}{d\mu}}}\\
&= -\log \crbk{\mu(A)}
\end{align*}
where the last inequality follows from the assumption that $\nu(A^c) = 0$. Since $\mu(A)\leq 1-\bar p_\wedge$, the above inequality implies that $\delta\geq \delta+\psi_\wedge$, which is a contradiction. 
\par This implies that the inequality constraint $\nu(s)\geq 0$ in $OPT_1$ is never active. Therefore, we can use the Lagrangian
\[
\cL(\nu,\lambda,\theta) = \nu\sqbk{\log \crbk{\frac{d\nu}{d\mu}}^2}-\lambda \nu\sqbk{\log \crbk{\frac{d\nu}{d\mu}}} +\lambda \delta - \theta\nu[1] + \theta.
\]
Observe that for any feasible $\nu$, $\del_\nu D_{\text{KL}}(\nu||\mu) = 1+\log(d\nu/d\mu)$, $\del_\nu \nu[1] = 1$ are never linearly dependent. So, any feasible point is a regular point of the equality constraints. Therefore by Chapter 11.3 in \cite{luenberger1984linear}, the KKT conditions are necessarily satisfied. 
\par We take derivative
\[
\del_{\nu}\cL(\nu,\lambda,\theta) = \log \crbk{\frac{d\nu}{d\mu}}^2 + 2\log \crbk{\frac{d\nu}{d\mu}} - \lambda \log \crbk{\frac{d\nu}{d\mu}} -\lambda - \theta.
\]
If we define $l = \log \crbk{\frac{d\nu}{d\mu}}$, the KKT condition implies that
\[
l^2+ (2- \lambda^*) l -\lambda^* - \theta^* = 0 \implies l_{\pm} = \frac{1}{2}\crbk{\lambda^*-2\pm\sqrt{(\lambda^*-2)^2 +4(\lambda^* + \theta^*) }}.
\]
Since $\nu[1] = 1$, we must have that $l_+\geq 0$ and $l_-\leq 0$. Define $S_+ = \set{s\in S:l(s)=l_+}$ We can write $\nu(s) = [e^{l_+}\1_{S_+}(s) + e^{l_-} \1_{S_+^c}(s)]\mu(s)$. Note that if $\mu(S_+) = 0,1$, then $\mu = \nu$ violating the constraint. So, $\mu(S_+) \neq 0,1$. Restrict $\mu,\nu$ on $\cG = \sigma(\set{S_+,S_-})$, then 
\[
l = \log\crbk{\frac{\nu(S_+)}{\mu(S_+)}}\1\set{s\in S_+}+ \log\crbk{\frac{\nu(S_+^c)}{\mu(S_+^c)}}\1\set{s\in S_+^c} = \log \crbk{\frac{d\nu|_\cG}{d\mu|_\cG}}.
\]
and
\[
OPT_1 = \nu|_\cG\sqbk{\log \crbk{\frac{d\nu|_\cG}{d\mu|_\cG}}^2}. 
\]
Also under this notation, 
\[
D_{\text{KL}}(\nu|_\cG ||\mu|_\cG) = l_+\nu(S_+)+ l_-\nu(S_-) = \nu\sqbk{\log\crbk{\frac{d\nu}{d\mu}}} = \delta. 
\]
Therefore, 
\[
OPT_1 \geq \inf_{\cG = \sigma(\set{A,A^c}):A\in \cS}\quad \inf_{\eta|_\cG :D_{\text{KL}}(\eta|_\cG||\mu|_\cG) = \delta} \eta|_{\cG}\sqbk{\log\crbk{\frac{d\eta|_\cG}{d\mu|_\cG}}^2} .
\] 
We define: 
\[
OPT_2:= \inf_{\cG = \sigma(\set{A,A^c}):A\in \cS}\inf_{\eta|_\cG :D_{\text{KL}}(\eta|_\cG||\mu|_\cG) = \delta} \eta|_{\cG}\sqbk{\log\crbk{\frac{d\eta|_\cG}{d\mu|_\cG}}^2} .
\]
As mentioned before, $A\neq S,\varnothing$ because of the constraint.
\par Next, we lower bound $OPT_2$. Notice that measureable strict subsets under $\cG$ is only $A$ and $A^c$. So, suppose $\eta|_{\cG}(A) = q$, we consider the following program
\begin{align*}
 \inf_{0< q <1} &\quad obj_2(q,b) := q \log\crbk{\frac{q}{b}}^2 + (1-q)\log\crbk{\frac{1-q}{1-b}}^2\\
 s.t. & \quad kl(q,b):=q \log\crbk{\frac{q}{b}} + (1-q)\log\crbk{\frac{1-q}{1-b}} = \delta
\end{align*}
where $b = \mu(S_+)$. This lower bounds $OPT_2$. The rest of the proof is denoted to compute the above program.

Note that w.l.o.g. we can assume $b \leq 1/2$ because if $b > 1/2$, we  change to new variable $b' = 1-b'< 1/2$ and $q' = 1-q$. Compute the second derivatives
\begin{align*}
d_qd_q kl(q,b) &= \frac{1}{q} + \frac{1}{1-q} > 0;\\
d_bd_b kl(q,b) &= \frac{q}{b^2} + \frac{1-q}{(1-b)^2} > 0;
\end{align*}
i.e. $kl(\cd,b),kl(q,\cd)$ are convex on $[0,1]$. So, its maximum is attained on the boundary: $kl(1,b) =-\log b$ and $kl(0,b) = -\log(1-b)$. By assumption, $1-e^{-\delta} < \bar p_\wedge\leq b = \mu(S_+)\leq 1-\bar p_\wedge < e^{-\delta}$. So, $kl(q,b) = \delta$ has 2 solutions $q_1< b < q_2$. Moreover
\begin{align*}
\log\crbk{\frac{q_i}{b}}-\log\crbk{\frac{1-q_i}{1-b}} &= \frac{1}{q_i}\crbk{kl(q_i,b) - \log\crbk{\frac{1-q_i}{1-b}}}= \frac{1}{q_i}\crbk{\delta +\log\crbk{\frac{1-b}{1-q_i}} }\\
&=\frac{1}{1-q_i}\crbk{ \log\crbk{\frac{q_i}{b}}-kl(q_i,b) }= -\frac{1}{1-q_i}\crbk{\delta +\log\crbk{\frac{b}{q_i}} }
\end{align*}
So, we have
\begin{align*}
OPT_2 -\delta^2  &\geq \inf_{i=1,2} obj_2(q_i) - kl(q_i,b)^2\\
&= \inf_{i=1,2}q_i(1-q_i)\crbk{\log\crbk{\frac{q_i}{b}}-\log\crbk{\frac{1-q_i}{1-b}}}^2\\
&= \inf_{i=1,2}\max\crbk{\frac{(1-q_i)}{q_i}\crbk{\delta +\log\crbk{\frac{1-b}{1-q_i}} }^2,\frac{q_i}{(1-q_i)}\crbk{\delta +\log\crbk{\frac{b}{q_i}} }^2}
\end{align*}
Observe that if $q > b$, then $1-q < 1-b$. So, 
\[
OPT_2-\delta^2\geq \delta^2\inf_{i=1,2}\frac{1-q_i}{q_i}\1\set{q> b}+\frac{q_i}{1-q_i}\1\set{q\leq b}.
\]
Now if we let $1/2 \geq b(\psi) = 1-e^{-\delta-\psi}$. By convexity and $kl \geq 0$ while $kl(b(\psi),b(\psi)) = 0$, $kl(\cd,b(\psi))$ is decreasing for $q \in[0, b]$. 
\begin{align*}
kl(q,b(\psi)) &= (1-q)(\delta+\psi) - q\log (1-e^{-\delta-\psi})+ q\log q + (1-q)\log(1-q)\\
&= (1-q)(\delta+\psi) +q\xi \log (\delta+\psi)+ q\log q + (1-q)\log(1-q)\\
\end{align*}
where we use $1-e^{-\delta-\psi} = e^{0}-e^{-\delta-\psi} = (\delta+\psi)e^{-\xi}$ for some $\xi \in (0,\delta+\psi)$. Also, $d_q (1-q)\log(1-q) = -1-\log(1-q)\geq -1$
\begin{align*}
kl(q,b)&\geq \delta + \psi +  q(\eta\log(\delta+\psi)-\delta-\psi - 1) +q\log q \\
&\geq \delta + \psi +  q((\delta+\psi)\log(\delta+\psi)-\delta-\psi - 1)+q\log q\\
&\geq \delta + \psi +  q(-1-\delta-\psi - 1)+q\log q\\
&\geq \delta + \psi -4 q +q\log q\end{align*}
We let $q(\psi) = -(\psi/c)/\log(\psi/c)$ for some $c > 1$. Note that when $\psi = 0$, $\psi -4 q(\psi) +q(\psi)\log q(\psi) = 0$.
We want $kl(q(\psi),b(\psi)) > \delta$; hence it suffices to show that
\[
d_{\psi}(\psi -4 q(\psi) +q(\psi)\log q(\psi))\geq 0.
\]
Let $\phi = -\log(\psi/c)$ We compute
\begin{align*}
d_\psi (\psi -4 q(\psi) +q(\psi)\log q(\psi)) &= \frac{\log \left(\frac{\psi }{c \phi }\right)}{c \phi ^2}-\frac{3}{c \phi ^2}+\frac{\log \left(\frac{\psi }{c \phi }\right)}{c \phi }-\frac{3}{c \phi }+1\\
&=\frac{1}{c \phi ^2}\crbk{c \phi ^2+(\phi +1) \log \left(\frac{\psi }{c \phi }\right)-3 \phi -3}\\
&=\frac{1}{c \phi ^2}\crbk{(c-1) \phi ^2-4 \phi -(\phi +1) \log (\phi )-3}
\end{align*}
Note that $c > e^{-1}$ and $\psi\in(0,1]$ implies that $\phi > 1$, $\log\phi\leq \phi-1$. Therefore,
\[
d_\psi (\psi -4 q +q\log q) \geq \frac{1}{c \phi ^2}\crbk{(c-1) \phi ^2-4 \phi -(\phi +1) (\phi -1)-3} = \frac{1}{c \phi ^2}\crbk{(c-2) \phi ^2-4 \phi-2}
\]
We see that if we choose $c = 8$, $d_\psi (\psi -4 q +q\log q) > 0$. Hence $kl(q(\psi),b(\psi)) \geq \delta$ for all $\psi$. Continuity and monotonicity imply that $q(\psi) < b(\psi)$ and there exists $q_1(\psi)\in [q(\psi), b(\psi)]$ s.t. $kl(q_1(\psi),b(\psi)) = \delta$. 

By the same argument, we can show that for $b'(\psi) = e^{-\delta-\psi}$, we have that $q_2(\psi)\in [ b'(\psi), 1-q(\psi)]$. 
Therefore, we conclude that for $1-e^{-\delta-\psi}\leq b \leq e^{-\delta-\psi}$
\[
OPT_2 - \delta^2\geq -\delta^2\frac{\psi/8}{\log(\psi/8)}.
\]
Recall that $\bar p_\wedge = 1-e^{-\delta - \psi_\wedge}$ and that $p_{\wedge}\leq \mu(S_+)\leq 1-\bar p_{\wedge}$ we conclude that
\begin{align*}
\mu'\sqbk{\log \crbk{\frac{d\mu'}{d\mu}}^2} - \delta^2 
&\geq \inf_{b\in[\bar p_\wedge,1-\bar p_\wedge]} OPT_2-\delta^2\\
&\geq  -\frac{\delta^2\psi_\wedge}{8\log(\psi_\wedge/8)}.
\end{align*}
\end{proof}

\subsubsection{Proof of Lemma \ref{lemma:alpha_ratio_bound}}
\begin{proof}
Recall that for $\omega\in\Omega_{s,a,n}(p)$, we have \eqref{eqn:Omega_meas_equiv}. Write
\begin{align*}
\sup_{\alpha\in K}\frac{\alpha m_{n}[w]^2}{\mu_{n}(t_3)[w]^2}&= \max\set{\sup_{\alpha\in [0,c \|u\|_\infty]}\frac{\alpha m_n[w]^2}{\mu_{n}(t_3)[w]^2} , \sup_{\alpha\in [c\|u\|_\infty,\delta\inv \|u\|_\infty]}\frac{\alpha  m_n[w]^2}{\mu_{n}(t_3)[w]^2}}\\
&=:\max\set{J_1(c),J_2(c)}
\end{align*}
We first bound $J_2(c)$
\[
J_2(c) = \sup_{\alpha\in [c\umax,\delta\inv \umax]}\frac{\alpha m_n[e^{- (u+\umax)/\alpha}]^2}{\mu_{n}(t_3)[e^{- (u+\umax)/\alpha}]^2}
\]
For simplicity, let $w':= e^{- (u+\umax)/\alpha}$. Recall that $m_n = \mu^O - \mu^E$, so $m_n[1] = 0$ and
\[
\begin{aligned}
\alpha m_n[e^{- (u+\umax)/\alpha} ]^2 & =  (m_m[\alpha^{1/2} (e^{- (u+\umax)/\alpha} -1)])^2.\\
\end{aligned}
\]
Define and note that $v := \alpha^{1/2}( e^{- (u+\umax)/\alpha} -1) < 0$. Then
\begin{align*}
\frac{\alpha m[w']^2}{\mu_{n}(t_3)[w']^2} &= \frac{m[v]^2}{\mu_{n}(t_3)[w']^2}\\
&= \frac{1}{\mu_{n}(t_3)[w']^2}\mu_n(t_3)\sqbk{\frac{dm_n}{d\mu_n(t_3)}v}^2\\
&\leq \frac{\mu_n(t_3)\sqbk{-v}^2}{\mu_{n}(t_3)[w']^2}\norm{\frac{dm_n}{d\mu_n(t_3)}}_{L^\infty(\mu)}^2\\
&\leq \norm{\frac{v}{w'}}_{L^\infty(\mu)}^2\norm{\frac{dm_n}{d\mu_n(t_3)}}_{L^\infty(\mu)}^2
\end{align*}
We defer the proof of the following claim:
\begin{lemma}\label{lemma:sup_|v/w|_infty_bound}
\[
\sup_{\alpha\in [c\umax,\delta\inv \umax]}\norm{\frac{v}{w'}}_{L^\infty(\mu)} \leq (c\umax)^{1/2}(e^{2/c}-1)
\]
\end{lemma}
Therefore, 
\[
J_2(c) \leq c\umax(e^{2/c}-1)^2\norm{\frac{dm_n}{d\mu_n(t_3)}}_{L^\infty(\mu)}^2.
\]
Assuming that $\delta\leq \log 2/2$, choose $c = 2/\log 2$
\begin{align*}
\sup_{\alpha\in K}\frac{\alpha m_{n}[w]^2}{\mu_{n}(t_3)[w]^2}
&=\max\set{J_1(c),J_2(c)}\\
&\leq \max \set{c\|u\|_\infty,c\umax(e^{2/c}-1)^2}\norm{\frac{dm_n}{d\mu_n(t_3)}}_{L^\infty(\mu)}^2\\
&\leq  3\umax \norm{\frac{dm_n}{d\mu_n(t_3)}}_{L^\infty(\mu)}^2
\end{align*}
which completes the proof.
\end{proof}
\subsubsection{Proof of Lemma \ref{lemma:sup_|v/w|_infty_bound}}
\begin{proof}
We bound
\[
\begin{aligned}
\norm{\frac{v}{w'}}_{L^\infty(\mu)} &= \esssup_{\mu} \alpha^{1/2}  ( e^{(u(s)+\umax)/\alpha} -1) \\
&\leq  \alpha^{1/2} (e^{2\umax/\alpha} -1)
\end{aligned}
\]

Compute derivative: let $\beta = 2\umax/\alpha$
\begin{align*}
\frac{d}{d\alpha}\alpha^{1/2} (e^{2\umax/\alpha} -1) &= \frac{e^{2 \umax/\alpha}-1}{2 \alpha^{1/2}}-\frac{2 \umax e^{2 \umax/\alpha}}{\alpha^{3/2}}\\
&= \frac{1}{2}(e^{\beta}(1 -2\beta)-1 
)\alpha^{-1/2}
\end{align*}
Notice that when $\beta = 0$, $e^{\beta}(1 -2\beta)-1 = 0$. Moreover, 
\[
\frac{d}{d\beta}e^{\beta}(1 -2\beta) = -e^\beta(1+2b) < 0;
\]
i.e. $e^{\beta}(1 -2\beta)$ decreasing. Therefore, for $\alpha > 0$
\[
\frac{d}{d\alpha}\alpha^{1/2} (e^{2\umax/\alpha} -1) < 0;
\]
i.e. $\alpha^{1/2} (e^{2\umax/\alpha} -1)$ is decreasing in $\alpha$. Hence
\[
\begin{aligned}
\sup_{\alpha\in [c\umax,\delta\inv \umax]}\norm{\frac{v}{w'}}_{L^\infty(\mu)}&\leq \sup_{\alpha\in [c\umax,\delta\inv \umax]}\alpha^{1/2} (e^{2\umax/\alpha} -1) \\
&= (c\umax)^{1/2}(e^{2/c}-1)
\end{aligned}
\]
establishing the claim. 
\end{proof}

\subsection{Proof of Lemma \ref{lemma:sub_g_4_max}}
\begin{proof}
For any $\lambda > 0$, consider an increasing function $\phi_\lambda(z) = \exp(\lambda z^{1/4})$ for $z\geq 0$. Since $Z\geq 0$, 
\[
\begin{aligned}
\phi_\lambda( EZ) &= \phi_\lambda( EZ\1\set{Z > (3/\lambda)^4}+ EZ\1\set{Z \leq (3/\lambda)^4})\\
&\leq \phi_\lambda( EZ\1\set{Z > (3/\lambda)^4}+ (3/\lambda)^4 P(Z \leq (3/\lambda)^4))\\
&\leq \phi_\lambda( EZ+ (3/\lambda)^4 )\\
\end{aligned}
\]
 By taking second derivatives, one can see that $\phi_\lambda(z)$ is convex for $z \geq  (3/\lambda)^4$. Therefore, by Jensen's inequality
\[
\begin{aligned}
\phi_\lambda( EZ)&\leq E\left[\phi_\lambda(Z+ (3/\lambda)^4 )\right]\\
&= e^3 E\left[\exp(\lambda\1\max_{i=1\ds n} |Y_i|)\right]\\
&\leq e^3 \sum_{i=1}^n Ee^{\lambda |Y_i|} 
\end{aligned}
\]
Since $\set{Y_i}$ are Sub-Gaussian,
\[
P(|Y_i|> t)\leq 2\exp\crbk{-\frac{t^2}{2\sigma^2}},
\]
which implies
\[
\log Ee^{\lambda |Y_i|}\leq 4\sigma^2\lambda^2.
\]
Therefore,
\[
\begin{aligned}
\log\phi_\lambda (EZ) &= \lambda \crbk{E\max_{i=1\ds n}Y_i^4}^{1/4}\\
&\leq 3+\log n + 4\sigma^2\lambda^2. 
\end{aligned}
\]
Rearrange and take infimum over $\lambda > 0$, we conclude
\[
\begin{aligned}
E\max_{i=1\ds n}Y_i^4&\leq \crbk{\inf_{\lambda > 0}\frac{3+\log n }{\lambda}+ 4\sigma^2\lambda}^4\\
&\leq 16 \sigma^4 \crbk{3+\log n}^2
\end{aligned}
\]
\end{proof}
\subsection{Proof of Lemma \ref{lemma:sup_f_bound}}
\begin{proof}
\[
\sup_{\alpha\geq 0} f(\nu, \alpha) \geq \lim_{\alpha\da 0}f(\nu,\alpha) = \essinf_{\nu}u \geq\essinf_{\mu}u\geq -\|u\|_{L^\infty(\mu)}
\]
On the other hand, since the $\sup$ is achieved on compact $K$. For optimal $\alpha_\nu^* > 0$, 
\[
\begin{aligned}
\sup_{\alpha\geq 0}f(\nu,\alpha)
&\leq \|u\|_{L^\infty(\nu)} -\alpha_\nu^*\log \nu[e^{-(u-\|u\|_{L^\infty(\nu)})/\alpha_\nu^*}]\\
&\leq \|u\|_{L^\infty(\mu)}
\end{aligned}
\]
where the last line follows from that $\nu[e^{-(u-\|u\|_{L^\infty(\nu)})/\alpha_\nu^*}] > 0$ and $ \nu\ll\mu$. Also, if $\alpha_\nu^* = 0$, the above holds trivially. 
\end{proof}

\section{Numerical Experiment}
\subsection{Test of Convergence on the Hard MDP}
For this numerical experiment using MDP \ref{fig:hard_mdp_instance}, we run the algorithm to produce independent 200 trajectories of 5000 iterations of Algorithm \ref{alg.Q_learning} under the rescaled linear and constant stepsize. Denote the estimated $Q$-function under rescaled linear ($\alpha_k = 1/(1+(1-\gamma)k)$) and constant stepsize ($\alpha = 0.008$) with $\hat Q$ and $\bar Q$ respectively. For each $\gamma = 0.7,0.8,0.9$, we produce $\set{\hat Q_{\delta,k}^{(i)}(\gamma),k=0,\ds,5000}$ and $\set{\bar Q_{\delta,k}^{(i)}(\gamma),k=0,\ds,5000}$ for $i = 1,2,\ds,200$. We use the size of the uncertainty set $\delta = 0.1$. 
\par Figure \ref{fig:hard_mdp_convergence} plots the lines that linearly interpolates
\[
\set{\crbk{\lg k , \lg\crbk{\frac{1}{200}\sum_{i=1}^{200} \norm{\hat Q_{\delta,k}^{(i) }(\gamma) - Q_\delta^*(\gamma)}_\infty}};  k= 0,1,\ds,5000}
\]
for $\gamma =  0.7,0.8,0.9$ as well as a line of slope $-1/2$ in the lg-lg scale as a reference. Figure \ref{fig:hard_mdp_convergence_const} plots the lines that linearly interpolates
\[
\set{\crbk{k , \lg\crbk{\frac{1}{200}\sum_{i=1}^{200} \norm{\bar Q_{\delta,k}^{(i) }(\gamma) - Q_\delta^*(\gamma)}_\infty}};  k= 0,1,\ds,5000}.
\]
\subsection{Test of Convergence for different $\delta$} 
\par
\begin{figure}[ht]
    \centering
    \includegraphics[width = 0.4\linewidth]{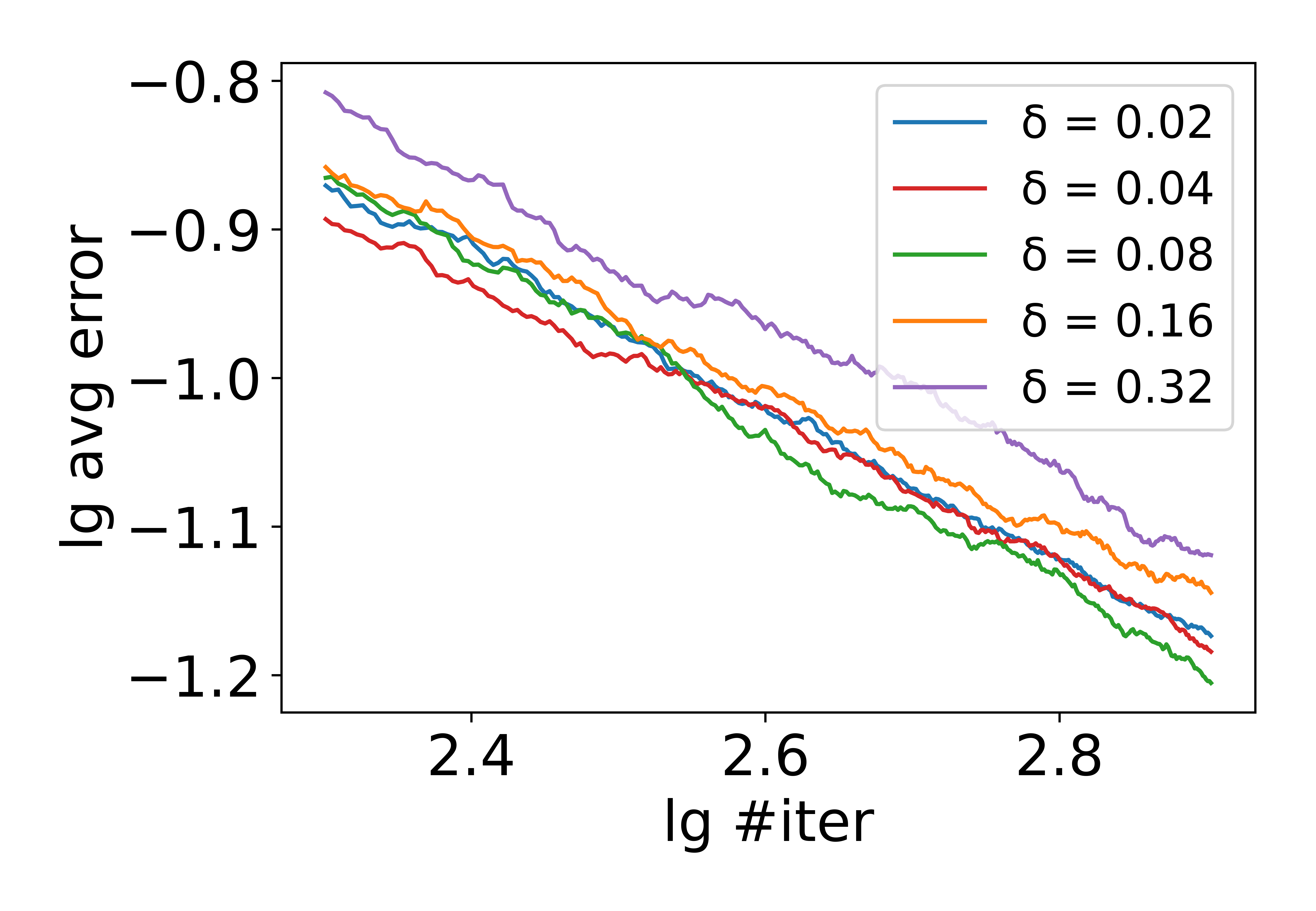}
    \includegraphics[width = 0.4\linewidth]{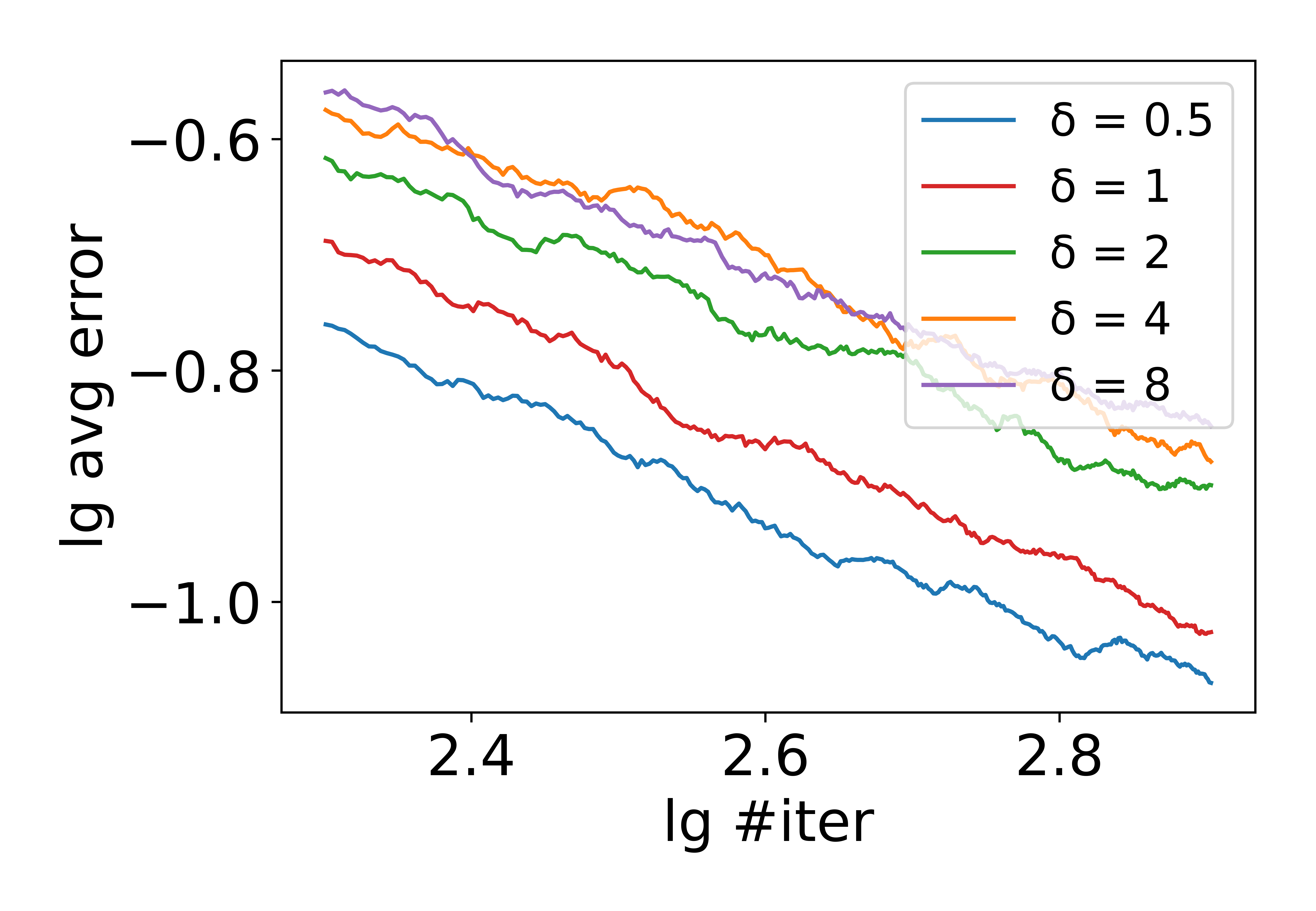}
    \caption{Test convergence for different $\delta$}
    \label{fig:test_delta}
\end{figure}

 We also numerically explore the complexity behavior when we change $\delta$. We use the same plotting procedure as in Figure \ref{fig:hard_mdp_convergence} only with different values of $\delta$. 
This give us Figure \ref{fig:test_delta}. We observe that indeed as $\delta\da 0$, the complexity is not sensitive to a change in $\delta$. This confirms our conjecture that the dependence on $\delta$ should be $O(1)$ as $\delta\da0$. Also interestingly, as we increase $\delta$, the complexity also becomes insensitive.
\subsection{Test of $\gamma$ Dependence on the Hard MDP} 
For Figure \ref{fig:hard_mdp_gamma_dep}, we run 200 trajectories Algorithm \ref{alg.Q_learning} with rescaled linear stepsize $\alpha_k = 1/(1+(1-\gamma)k)$ for 10 evenly spaced $\gamma_j \in[0.7, 0.95], \gamma_1 = 0.7,\gamma_{10} = 0.95$ until a fixed iteration $k = 500, 1000,1500$. For each This produce a data set $\set{\hat Q_{\delta,k}^{(i)}(\gamma_j); i = 1,2\ds,200;j = 1,2,\ds,10}$ for $k = 500, 1000,1500$. We still use the size of the uncertainty set $\delta = 0.1$. Figure \ref{fig:hard_mdp_gamma_dep} plots the scattered pairs
\[
\set{\crbk{\lg(1-\gamma_j), \lg \crbk{\frac{1}{200}\sum_{i=1}^{200} \norm{\hat Q_{\delta,k}^{(i)}(\gamma_j) - Q_\delta^*(\gamma_j)}_\infty}};j = 1,2,\ds,10}
\]
and the least square regression line for each $k = 500, 1000,1500$.

\subsection{Lost-Sale Inventory Control Model}
In this experiment, we use $\delta = 0.5$, $\gamma = 0.7$. We run 300 trajectories of 5000 iterations of our Algorithm  \ref{alg.Q_learning} ($g = 5/8$) and that in \cite{pmlr-v162-liu22a} ($g=0.499$). In both cases we use the rescaled-linear stepsize. We also record the number of sample used by trajectory $i$ at iteration $k$ (denote by $\hat n_{g,k}^{(i)}$) This produces data $\set{\hat n_{g,k}^{(i)}, \hat Q_{\delta,g,k}^{(i)}; i=1,\ds,300; k = 0,\ds,5000}$ for $g = 5/8$ and $g = 0.499$. 

\par Figure \ref{fig:inventory_avg_avg}  plots the linear interpolation of points
\[
\set{\crbk{\lg \crbk{\frac{1}{300}\sum_{i=1}^{300} \hat n_{g,k}^{(i)}} , \lg\crbk{\frac{1}{300}\sum_{i=1}^{300} \norm{\hat Q_{\delta,k}^{(i) }(\gamma) - Q_\delta^*(\gamma)}_\infty}};  k= 1000,1001,\ds,5000}
\]
for $g = 5/8, 0.499$. For presentation clearness, Figure \ref{fig:inventory_sample_at_err} plots the \textit{smoothed} scattering of the data
\[
\set{\crbk{ \hat n_{g,k}^{(i)}, \lg\crbk{\norm{\hat Q_{\delta,k}^{(i) }(\gamma) - Q_\delta^*(\gamma)}_\infty}};  i = 1,\ds,300;k= 1000,\ds,5000}
\]
The smoothing is in error, over a window of size $w =0.0001$; i.e. if we define
\[
\bar n_{g,k}^{(i)}(w) := \text{mean}\set{n_{g,m}^{(n)}: 0 \leq \hat Q_{\delta,g,m}^{(n)}- Q_{\delta,g,k}^{(i)}\leq w },
\]
then Figure \ref{fig:inventory_sample_at_err} plots the linear interpolation of points
\[
\set{\crbk{\hat Q_{\delta,g,k}^{(i)},\bar n_{g,k}^{(i)}(w)};i = 1,\ds,300; k = 1000,\ds,5000}
\]for $g = 5/8, 0.499$.

\end{document}